\documentclass[lettersize,journal]{IEEEtran} 
\usepackage{amsmath,amsfonts}
\usepackage{amssymb}
\usepackage{bm}
\usepackage{algorithmic}
\usepackage{algorithm}
\usepackage{array}
\usepackage{textcomp}   
\usepackage{stfloats}
\usepackage{url}
\usepackage{verbatim}
\usepackage{graphicx}
\usepackage{cite}
\usepackage{xcolor}

\usepackage{amsthm}

\usepackage{cancel}
\usepackage[caption=false, font=footnotesize]{subfig}

\usepackage[normalem]{ulem}

\usepackage{empheq}  

\allowdisplaybreaks[4]

\usepackage{xpatch}

\usepackage{booktabs}

\begin{document}

\title{Agentic AI for Bilevel Long-Term Optimization of Policy-Driven Physical Layer Systems}

\author{Bingnan~Xiao,
        Chenhao~Yang, 
        Wei~Ni,~\IEEEmembership{Fellow,~IEEE},
        Xin~Wang,~\IEEEmembership{Fellow,~IEEE},
        and Tony~Q.~S.~Quek,~\IEEEmembership{Fellow,~IEEE}
\thanks{B. Xiao and X. Wang are with the Key Laboratory of EMW Information (MoE), College of Future Information Technology, Fudan University, Shanghai 200433, China (e-mail: 22110720061@m.fudan.edu.cn, xwang11@fudan.edu.cn).

C.~Yang is with the James Watt School
of Engineering, University of Glasgow, Glasgow G12 8QQ, U.K. (email:3165453Y@student.gla.ac.uk).

W. Ni is with the School of Engineering, Edith Cowan University, Perth, WA 6027, Australia (e-mail: wei.ni@ieee.org).

T. Q. S. Quek is with the Information Systems Technology and Design Pillar, Singapore University of Technology and Design, Singapore 487372 (e-mail: tonyquek@sutd.edu.sg).
}
}



\maketitle

\begin{abstract}
Network operators' changing policies, service requirements, and stringent real-time constraints render existing methods designed with fixed objectives and constraints ineffective. This paper presents Agentic long-term performance optimization (Agentic-LTPO), a nested bilevel optimization framework that can be applied to adaptive physical layer problem configuration. The key idea is to employ agentic AI to generate upper-level configurations in a bilevel optimization structure, where evolving operator policies, environment summaries, and historical experiences are translated into structured lower-level optimization problem configurations. The lower level solves the problems with updated configurations for real-time physical-layer decisions. Considering cell-free MIMO beamforming as a use case, we embody Agentic-LTPO by designing a new multi-agent decision process with retrieval-augmented experience-based verification in the upper level, together with a closed-form beamformer in the lower level. Experiments demonstrate that Agentic-LTPO exhibits strong adaptability to dynamic operator policies and effectively enhances the system’s long-term performance by 57.2\% compared to traditional methods.

\end{abstract}


\section{Introduction}\label{sec:intro}

Future wireless networks will be featured by dense connectivity, diverse service requirements, and software-defined operations~\cite{ngo2017cellfree}. The widespread deployment of coordinated physical-layer architectures, e.g., cell-free (CF) massive MIMO, gives rise to new beamforming and radio resource allocation designs necessitating distributed access points (APs) to jointly optimize transmission decisions under coupled quality-of-service (QoS) and power constraints~\cite{ngo2017cellfree,shi2011wmmse,Femenias2025from}. In a classical physical-layer control framework, the network operator first specifies an objective and constraints, after which a model-based solver produces decisions~\cite{shi2011wmmse}. In practice, operator policies, intents, and key performance indicators (KPIs) can change over time. Preferably, the physical layer would adapt to a policy-driven, non-stationary environment with time-varying objectives and constraints.

Deep learning (DL) and deep reinforcement learning (DRL) have helped relax the need for persistent optimization configurations of wireless control~\cite{liang2020deep}. DRL has been employed to learn 
\textcolor{black}{resource allocation and beamforming} policies from interactions with the environment, reducing dependence on explicit analytical models at runtime~\cite{cai2024deep,wang2024joint}. 
Learning-based methods can improve online adaptivity and generate low-latency control decisions for dynamic wireless environments~\cite{liang2020deep}. 
Unfortunately, these methods are designed with a pre-specified utility or reward function concerning a given series of KPIs. Once the operator's intent changes, e.g., from throughput to energy efficiency, the deployed learning rule or reward design needs to be modified, and the controller needs to be retrained. 

Recent advances in large language models (LLMs) and agentic AI offer a new possibility for and adaptive wireless control~\cite{wang2025Hierarchical,Nahum2026Intent,hang2024large,liang2026large,navidan2026toward}. Existing studies show that LLMs can interpret natural-language intents, retrieve relevant evidence, and coordinate multiple tools for networking tasks~\cite{wang2025Hierarchical,Nahum2026Intent,hang2024large}. 
In fact, most existing works have focused on intent extraction, configuration assistance, or control-plane orchestration~\cite{liang2026large,navidan2026toward}. Directly applying an LLM to generate beamformers is inappropriate since these actions must satisfy strict real-time and numerical feasibility requirements under instantaneous channel state information (CSI). 

A more appropriate use case of agentic AI is to configure the fast timescale physical layer controllers, e.g., beamforming optimizers, by interpreting operator policies, summarizing network behavior, and reusing historical experience at a slower timescale.
Following a bilevel optimization structure, this can leverage agentic AI's capability of interpreting natural-language policies, reasoning about long-term network behavior, and retrieving relevant historical configurations at the upper level to produce structured lower-level problem configuration parameters that reflect the operator's evolving policies and intents. It is non-trivial to establish a reliable interface between heterogeneous upper-level policy and intent inputs, and executable lower-level configuration parameters. The reason is that an AI agent must produce a structured and feasible configuration at the upper level, while the quality of the configuration can only be assessed indirectly through the lower-level responses accumulated over time.

\subsection{Related Works}
\label{sec:related_work}
\subsubsection{Physical-Layer Configuration for Coordinated Wireless Systems}
Significant effort has been devoted to physical-layer optimization for coordinated wireless systems. Multi-cell cooperative transmission and coordinated beamforming were investigated in~\cite{gesbert2010multi,shi2011wmmse} to mitigate inter-cell interference and improve network utility. Massive MIMO was studied in~\cite{marzetta2010noncooperative,larsson2014massive} to provide a scalable architecture for high spectral efficiency. Built on these advances, CF massive MIMO was introduced in~\cite{ngo2017cellfree}, and studied from the perspectives of precoding and power control~\cite{nayebi2017precoding}, user-centric design and implementation~\cite{interdonato2019ubiquitous,bjornson2020making}, and local/distributed processing~\cite{interdonato2020localpzf}. These works have typically assumed that the lower-level objective and constraints are specified beforehand, and cannot support time-varying beamforming problem settings with time-varying operator policies, operating rules, and KPIs.

\subsubsection{Learning-Based Wireless Control and Optimization Acceleration}
Recent advances in learning-based wireless control have reduced online complexity and improved adaptivity in dynamic environments. The growing roles of DL and DRL in wireless communications and networking have been articulated in~\cite {liang2020deep,mao2018deeplearning}. Representative DRL-based designs have been developed for dynamic multichannel access~\cite{wang2018dynamicmultichannel}, \textcolor{black}{online resource allocation~\cite{cai2024deep}, and dynamic beamforming design~\cite{wang2024joint}}, typically relying on general-purpose DRL frameworks. 

A large body of work has aimed to learn or accelerate structured optimization. Deep neural networks were trained for wireless resource management in~\cite{sun2017learn2opt}, sample-efficient optimization with limited supervision was investigated in~\cite{shen2020lorm}, and model-driven DL for physical-layer communications was reviewed in~\cite{he2019modeldriven}. Recently, deep unfolding and graph-based unrolling have been applied to weighted minimum mean-square error (WMMSE)-type wireless operation optimization, including matrix-inverse-free unfolding~\cite{pellaco2022matrixinversefree}, GNN-assisted WMMSE unrolling for power allocation~\cite{chowdhury2021uwmmse}, deep graph unfolding for beamforming~\cite{chowdhury2024deepgraph}, and knowledge-driven WMMSE-unrolled resource allocation~\cite{yang2024uwgnn}. Compared with pure model-based optimization, these methods improve decision latency, approximation capability, and online deployability, especially when the optimization structure is known and stable. Yet, they have been designed largely around pre-specified utilities, rewards, or optimization templates. They do not address how to interpret evolving policy intents and convert them into structured physical-layer problem configurations. 

\subsubsection{Intent-Aware Networking and Agentic Wireless Control}
A small but rapidly growing body of works has connected intent understanding, LLMs, and agentic AI with communication networks. Early intent-based networking studies established the concepts, abstractions, and service-assurance mechanisms of intent-driven control~\cite{pang2020intentdriven,leivadeas2023ibn}, while LLM-assisted intent extraction for 5G network management was explored in~\cite{manias2024intent5g}. These studies suggest that high-level operator intent can be translated into structured network-side semantics. Recent works have increasingly shifted attention from intent parsing to broader LLM-empowered network intelligence~\cite{hang2024large,liang2026large}.
LLM-based in-context learning and optimization were explored for power control and resource allocation~\cite{zhou2024iclwireless,peng2025llmoptira}. 
LLM-assisted algorithm generation was explored for networking~\cite{he2024nada}, and generic tool-using and multi-agent LLM paradigms were developed in~\cite{yao2023react,schick2023toolformer,wu2024autogen}, offering new building blocks for reasoning, tool invocation, and hierarchical autonomous control. 

Recent studies have also instigated agentic AI for open and intelligent RANs, including tool-oriented agentic communications, autonomous control architectures for open 6G networks~\cite{elkael2025agentran}, multi-scale agentic control and management for O-RAN~\cite{navidan2026oran}, conflict-aware multi-agentic rApp policy orchestration~\cite{li2026multiagentic}, and intent-driven optimization for CF O-RAN~\cite{shokouhi2026intentoran}. These works have focused on architectural autonomy, orchestration, intent translation, or direct LLM-assisted optimization. It is non-trivial to enable reasoning-capable agents to reliably translate evolving high-level intents into configurations of structured physical-layer controllers when the underlying objectives and constraints vary over time.

\subsection{Contributions}\label{sec:contributions}

This paper proposes a new framework, named Agentic long-term performance optimization (Agentic-LTPO), to adaptively update optimization configuration specified for effective control of wireless physical layer, where network operator's intents, policies, operating rules, and KPIs can change over time. The key contributions are summarized as follows:
\begin{itemize}

\item
  We formulate adaptive physical-layer control under changing operator's policies and intents as a nested bilevel problem, where the upper and lower levels manage policy-driven optimization configuration and instant physical layer optimization, respectively. Agentic-LTPO decouples these levels into two timescales, allowing agentic AI to interpret the operator's intents at a large timescale while preserving the optimality of lower-level optimization at the lower level.

  \item
  Considering CF-MIMO beamforming to showcase Agentic-LTPO, we design a multi-agent collaboration architecture for the upper level to convert policy inputs, environment summaries, and historical experience into physically feasible configuration parameters. This is achieved through four agentic roles, including interpretation, observation, planning, and criticism, as well as a planner--critic refinement loop.

  \item
    To anchor the upper-level decisions on accumulated operating evidence (as opposed to isolated generation), we design a retrieval augmented generation (RAG) module that maintains a policy memory and a case memory to supply relevant empirical evidence during policy interpretation and configuration evaluation.

  \item
    We reveal that the robust energy-minimization problem of low-level beamforming admits a closed-form worst-case signal-to-interference-plus-noise ratio (SINR) bound under a zero-forcing criterion, leading to an efficient per-slot solver with global optimality and linear complexity for evaluating configurations generated by the upper level.




\end{itemize}
Extensive experiments are conducted on CF-MIMO beamforming systems under both random and piecewise-stationary operator policy settings. Agentic-LTPO improves the cumulative communication utility by $57.2\%$ over the static baseline, confirming the benefit of adaptively updating lower-level configurations under evolving operator policies. 
We also examine the KPI responses and configuration trajectories across different policy regimes, and compare raw natural-language policies with oracle structured-policy inputs to evaluate the sensitivity of upper-level decisions to language grounding. 
It is demonstrated that Agentic-LTPO translates policy regimes into interpretable configuration updates and target-KPI responses while reducing the impact of language ambiguity.

The rest of this paper is organized as follows. Section~II introduces the system model. Section~III formulates the two-timescale Agentic-LTPO problem and proposes the new upper-level multi-agent collaboration mechanism, the lower-level optimization method, and the overall algorithm implementation. Section~IV provides the experimental results, followed by the conclusions in Section~V.
\begin{table}[!t]
\centering
\caption{Notation and definitions.}
\label{tab:notation_definitions}
\footnotesize
\setlength{\tabcolsep}{1.5pt}
\renewcommand{\arraystretch}{1.08}
\begin{tabular}{@{}l@{\hspace{4pt}}p{0.72\columnwidth}@{}}\toprule
\textbf{Notation} & \textbf{Definition} \\
\midrule
$\mathcal{L}$, $L$ & Set and number of distributed APs \\
$\mathcal{K}$, $K$ & Set and number of users \\
$\mathcal{T}$, $T$ & Set and number of time slots \\
$N$, $T_s$ & Number of large time intervals and slots per interval \\
$\mathcal{T}^{(n)}$ & Slot set of interval $n$ \\
$\mathbf{h}_{\ell k}^{t}$, $\widehat{\mathbf{h}}_{\ell k}^{t}$ & Channel and its estimate from AP $\ell$ to user $k$ in slot $t$ \\
$\Delta\mathbf{h}_{\ell k}^{t}$, $\delta_{\ell k}^{t}$ & CSI error and its uncertainty radius \\
$\mathbf{w}_{\ell k}^{t}$, $\mathbf{W}^{t}$ & Beamforming vector and matrix in slot $t$ \\
$P_\ell^t$, $P_\ell^{\max}$ & Transmit power of AP $\ell$ in slot $t$ and power budget \\
$\mathrm{SINR}_{k}^{t,\mathrm{wc}}$ & Worst-case robust SINR of user $k$ in slot $t$ \\
$\Gamma_k^{(n)}$ & Target robust QoS level of user $k$ in interval $n$ \\
$\phi_\ell^{(n)}$ & AP power-budget exposure factor of AP $\ell$ in interval $n$ \\
$\boldsymbol{c}^{(n)}$ & Upper-level configuration vector for interval $n$ \\
$\boldsymbol{c}^{(n),r}$ & Candidate configuration in refinement round $r$ of interval $n$ \\
$\boldsymbol{d}^{(n),r}$ & Planner-generated adjustment in refinement round $r$ of interval $n$ \\
$a^{(n),r}$ & Acceptance flag returned by the critic in refinement round $r$ of interval $n$ \\
$P^{(n)}$, $\bar P^{(n)}$ & Operator policy profile and structured policy context in interval $n$ \\
$\mathcal{C}(\bar P^{(n)})$ & Feasible configuration set induced by $\bar P^{(n)}$ \\
$A_{\mathrm{int}}$, $A_{\mathrm{obs}}$ & Policy interpreter and network observer agents \\
$A_{\mathrm{plan}}$, $A_{\mathrm{crit}}$ & Configuration planner and performance critic agents \\
$S_{\mathrm{env}}^{(n)}$ & Environment summary used by the upper-level agents \\
$\mathbf{s}_{\mathrm{op}}^{(n)}$, $\mathbf{s}_{\mathrm{diag}}^{(n)}$ & Operational statistics and semantic diagnosis in $S_{\mathrm{env}}^{(n)}$ \\
$\mathcal{D}^{(n)}$ & Per-slot operating records collected over interval $n$ \\
$\mathbf{u}^t$ & Per-slot KPI measurements after deploying configuration \\
$\mathcal{E}^{(n)}$ & Cross-timescale experience buffer maintained by the upper level \\
$\mathcal{M}_{\mathrm{pol}}^{(n)}$, $\mathcal{M}_{\mathrm{case}}^{(n)}$ & Policy memory and case memory in the RAG module \\
$\mathcal{Z}_{\mathrm{pol}}^{(n)}$, $\mathcal{Z}_{\mathrm{case}}^{(n),r}$ & Retrieved policy evidence and case evidence \\
$R^t$, $\mathrm{EE}^t$ & Sum-rate and energy-efficiency metrics in slot $t$ \\
$\mathrm{KPI}^{(n)}$ & Interval-level KPI collection in interval $n$ \\
$J_{\mathrm{ag}}^{(n)}$ & Inference-energy proxy of Agentic AI in interval $n$ \\
$\lambda_R$, $\lambda_{\mathrm{EE}}$, $\lambda_J$ & Weights of sum-rate, energy efficiency, and agentic cost in the upper-level utility \\
$R_{\mathrm{ref}}$, $\mathrm{EE}_{\mathrm{ref}}$, $J_{\mathrm{ref}}$ & Normalization constants for the upper-level utility \\
$G(\mathrm{KPI}^{(n)})$ & Upper-level interval utility evaluated from communication KPIs and agentic cost \\
\bottomrule
\end{tabular}
\end{table}

\textit{Notation:}
Lower-case letters indicate scalars (e.g., $x$), boldface lower-case letters indicate vectors (e.g., $\mathbf{x}$), and boldface upper-case letters indicate matrices (e.g., $\mathbf{X}$). 
Calligraphic letters represent sets (e.g., $\mathcal{K}$, $\mathcal{L}$). 
$|\mathcal{S}|$ denotes the cardinality of a finite set $\mathcal{S}$. 
$[\mathbf{x}]_i$ denotes the $i$-th entry of a vector $\mathbf{x}$, and $[\mathbf{X}]_{i,j}$ denotes the $(i,j)$-th entry of a matrix $\mathbf{X}$. 
$(\cdot)^{H}$ denotes its Hermitian transpose. 
$\|\mathbf{x}\|_2$ denotes the Euclidean norm of a vector $\mathbf{x}$, and $\|\mathbf{X}\|_F$ denotes the Frobenius norm of a matrix $\mathbf{X}$. 
$\mathrm{diag}(\mathbf{x})$ denotes a diagonal matrix with diagonal elements given by vector $\mathbf{x}$.

\section{System Model}\label{sec:system_model}


This section introduces the Agentic-LTPO framework and its CF-MIMO downlink beamforming example.

\begin{figure}[!t]
	\centering
	\includegraphics[width=0.95\linewidth]{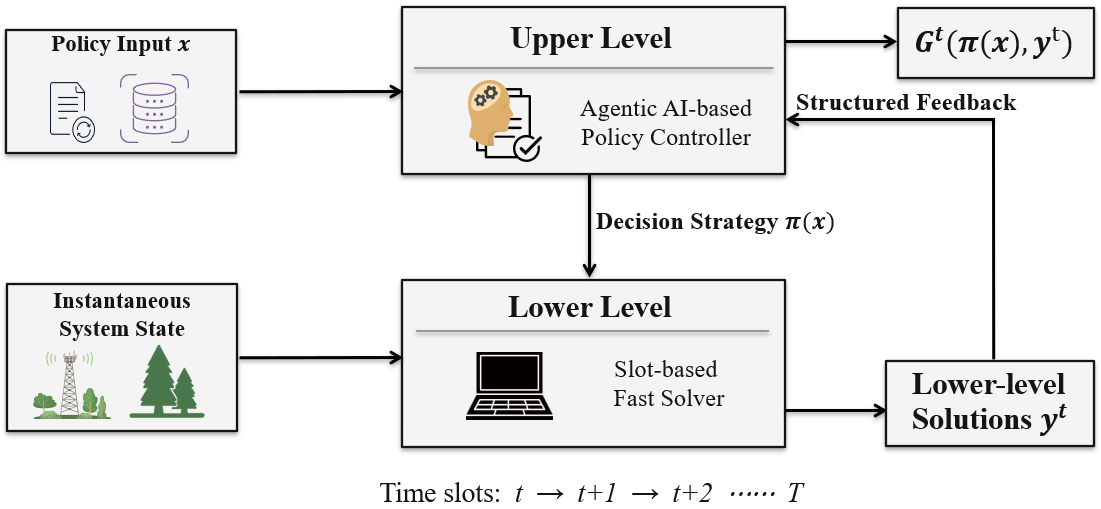}
	\caption{An illustration of the proposed Agentic-LTPO framework.}
	\label{fig:sys model}
\end{figure} 

\subsection{Agentic-LTPO Framework}
Agentic-LTPO admits a nested optimization framework, as depicted in Fig. 1.
To rapidly adapt to operator policies and channels and make effective use of the autonomous decision-making and reasoning abilities of agentic AI, the nested learning problem is formulated as
\begin{align} \label{obj func}
    \min _{\boldsymbol{\pi}(\boldsymbol{x})}  F(\boldsymbol{\pi}(\boldsymbol{x}),& \boldsymbol{y}^{1:T}) = \frac{1}{T} \sum_{t=1}^{T} w^t G^t( \boldsymbol{\pi}(\boldsymbol{x}),\boldsymbol{y}^t)
    \nonumber \\    
    \text { s.t. }  \boldsymbol{y}^t \! &\in \! {\operatorname{argmin}} _{\boldsymbol{y} \in \mathcal{Y}} f^t(\boldsymbol{\pi}(\boldsymbol{x}),\boldsymbol{y}), \forall t \!= \! 1,\cdots,T,   \nonumber \\  
    & g_i^t(\!\boldsymbol{\pi}(\boldsymbol{x})\!) \!\leq\! 0, h_i^t(\!\boldsymbol{\pi}(\boldsymbol{x}),\boldsymbol{y}\!) \!\leq\! 0, \forall i \!= \! 1,\cdots,I
\end{align}
where $\boldsymbol{x} \in \mathbb{R}^{d_1}$ denotes the input data evaluated under the decision policy $\boldsymbol{\pi}(\cdot)$ generated by agentic AI; $\boldsymbol{\pi}(\boldsymbol{x}) \in \mathbb{R}^{d_2}$ and $\boldsymbol{y} \in \mathbb{R}^{d_3}$ are the upper- and lower-level decision variables, respectively; $\boldsymbol{y}^{1:T}$ denotes the sequence of instantaneous lower-level responses across a time horizon of $T$ time slots, with the weight $w^t$ reflecting the long-term priority of slot $t$; each time slot $t$ has its specific upper- and lower-level objective functions, $G^t$ and $f^t$, driven by rapidly varying communication demands; 
$g_i^t$ defines the feasible boundary of the high-level policy space, and $h_i^t$ encapsulates the physical constraints accounting for operator policies and coupling Agentic AI's long-term intents with the instantaneous actions.

Problem~\eqref{obj func} provides an interface between a policy-driven upper level and a latency-sensitive lower level. 
The upper-level decision strategy $\boldsymbol{\pi}(\cdot)$ abstracts the long-term intent and configuration of a communication task (see Section III-A), and $\boldsymbol{\pi}(\boldsymbol{x})$ is updated based on aggregated observations at a large timescale interval. 
The lower level computes the instantaneous response $\boldsymbol{y}^t$ by performing a per-slot optimization subject to instantaneous constraints $g_i^t(\boldsymbol{\pi}(\boldsymbol{x}))$ and $h_i^t(\boldsymbol{\pi}(\boldsymbol{x}),\boldsymbol{y})$.\footnote{The nested problem construction of \eqref{obj func} is motivated by the stringent real-time requirements of physical-layer tasks: The per-slot decision $\boldsymbol{y}^t$ must be produced slot by slot, whereas the decision policy $\boldsymbol{\pi}(\boldsymbol{x})$ is executed at the large timescale spanning many slots.} 

To solve \eqref{obj func}, in slot $t$, the following steps are executed:
\begin{itemize}
    \item Upper-level configuration based on Agentic AI: 
    The upper level ingests the current policy profile $P$ from the operator and long-term performance summaries, input data $\boldsymbol{x}$, and generates the decision strategy $\boldsymbol{\pi}(\boldsymbol{x})$, which is broadcast to the lower level.

    \item Lower-level optimization based on fast solvers: 
    Conditioned on $\boldsymbol{\pi}(\boldsymbol{x})$, the lower level computes $\boldsymbol{y}^t\in\arg\min_{\boldsymbol{y}\in\mathcal{Y}} f^t(\boldsymbol{\pi}(\boldsymbol{x}),\boldsymbol{y})$ subject to the system constraints with a designed fast solver to meet the per-slot complexity and latency requirements.

    \item Upper-level update based on lower-level feedback:
    After lower-level optimization, structured feedback based on $\boldsymbol{y}^t$ is returned to the upper level to update its decision strategy $\boldsymbol{x}$ to compute $G^t\left(\boldsymbol{\pi}(\boldsymbol{x}), \boldsymbol{y}^t\right)$.     
\end{itemize}
Then, the next ($t+1$)-th time slot starts. This repeats until the expiration of the time horizon $T$.

\subsection{CF-MIMO Downlink Beamforming}
An embodiment of the Agentic-LTPO framework is instantiated in a CF-MIMO scenario, which features a two-timescale nested structure: 
Lower-level beamforming must be optimized at each slot based on instantaneous (often imperfect) CSI, while the upper-level configuration, including the target user QoS and AP power budgets, is adapted to operator policies and KPIs that evolve at a large timescale. 
The beamforming vectors across all AP-user pairs are coupled through inter-user interference, per-AP power constraints, and policy-dependent QoS requirements.

In the downlink CF-MIMO system, a set of $L$ distributed APs, $\mathcal{L}\triangleq\{1,\ldots,L\}$, jointly serve a set of $K$ single-antenna users, $\mathcal{K}\triangleq\{1,\ldots,K\}$. Each AP $\ell\in\mathcal{L}$ is equipped with $M$ antennas.
All APs are connected to a CPU through fronthaul with abundant bandwidth, centralized coordination~\cite{ngo2017cellfree,bjornson2020making}. Agentic AI is deployed at the CPU side, e.g., on an edge/cloud platform co-located with the CPU. It ingests long-term KPIs and policy inputs from the operator at the upper level, and updates the configuration passed to the lower-level beamforming solvers at the APs.


As described in Section II-A, the system operates for $T$ time slots, denoted as $\mathcal{T} = \{1,  \ldots, T\}$, which constitute $N$ upper-level decision intervals. Consider a block fading channel.
The channel remains unchanged in a time slot and can vary independently across slots. Let $\mathbf{h}_{\ell k}^t\in\mathbb{C}^{M}$ be the channel from AP $\ell$ to user $k$ in slot $t$. 
The communication symbol $s_k^t$ destined for user $k$ with $\mathbb{E}[|s_k^t|^2]=1$ is precoded by the beamforming vector $\mathbf{w}_{\ell k}^t\in\mathbb{C}^{M}$ at AP $\ell$, as given by
\begin{align}
    \mathbf{x}_\ell^t = \sum_{k=1}^K  \mathbf{w}_{\ell k}^t s_k^t, \ \forall t \in \mathcal{T}.
\end{align}

At each time slot $t$, we design the downlink beamforming matrix $\mathbf{W}^{t} := \big[\mathbf{w}_{1}^{t},\ldots,\mathbf{w}_{K}^{t}\big]\in\mathbb{C}^{LM\times K}$ with $\mathbf{w}_{k}^{t}\triangleq
\big[(\mathbf{w}_{1k}^{t})^{\intercal},\ldots,(\mathbf{w}_{Lk}^{t})^{\intercal}\big]^{\intercal}$ to meet the long-term objective dictated by the upper-layer policy specified by Agentic AI. The beamforming decisions are optimized based on instantaneous CSI, under the guidance of the slowly varying KPIs and operating policies.

\subsection{System Performance Metrics}

We introduce the performance metrics of tasks to support the subsequent nested optimization problem formulation.

\subsubsection{Quality of service}

In the downlink CF-MIMO system, the received signal of user $k$ at time slot $t$ is given by 
\begin{align}
y_k^t
= \underbrace{\left(\sum_{\ell=1}^{L} (\mathbf{h}_{\ell k}^t)^{H}\mathbf{w}_{\ell k}^t\right)s_k^t}_{\text{desired}}
+ \underbrace{\sum_{j\neq k}\left(\sum_{\ell=1}^{L} (\mathbf{h}_{\ell k}^t)^{H}\mathbf{w}_{\ell j}^t\right)s_j^t}_{\text{multiuser interference}}
+ n_k^t ,
\end{align}
where $n_k^t\sim\mathcal{CN}(0,(\sigma_k^t)^2)$ is the additive circularly symmetric complex Gaussian (CSCG) noise at user $k$ in slot $t$, with variance $(\sigma_k^t)^2$. Then, the SINR of user $k$ at slot $t$ is
\begin{align} \label{eq:sinr}
\mathrm{SINR}_k^t
= \frac{\left|\sum_{\ell=1}^{L} \left(\mathbf{h}_{\ell k}^t\right)^{H}\mathbf{w}_{\ell k}^t\right|^{2}}
{\sum_{j\neq k}\left|\sum_{\ell=1}^{L} \left(\mathbf{h}_{\ell k}^t\right)^{H}\mathbf{w}_{\ell j}^t\right|^{2} + (\sigma_k^t)^2 } .
\end{align}
We can use SINR as a performance metric to measure the QoS of a user in the system.

\subsubsection{Sum-rate}
The system can be subject to a network-level performance requirement. We characterize such a requirement using the sum-rate, defined as: At slot $t$,
\begin{align}  \label{eq:wsr}
R^{t}
= \sum_{k=1}^{K}  \log_2\!\left(1+\mathrm{SINR}_k^t\right).
\end{align}

\subsubsection{Total transmit power} At time slot $t$, the total transmit power of the downlink CF-MIMO system is defined as the total transmit power of all APs to all users, i.e.,
\begin{equation}
P_{\mathrm{tot}}^{t} = \sum_{\ell=1}^{L}\sum_{k=1}^{K}\big\|{\bf w}_{\ell k}^{t}\big\|_{2}^{2}.
\label{eq:ptot}
\end{equation}

\subsubsection{Energy efficiency} The energy efficiency of the system in time slot $t$ is defined as the ratio between the achieved sum-rate and the total transmit power, i.e.,
\begin{equation}
\mathrm{EE}^{t} = \frac{R^{t}}{P_{\mathrm{tot}}^{t}},
\label{eq:ee}
\end{equation}
which quantifies how efficiently the transmit power is converted to the throughput of the system.

\subsubsection{Agentic AI energy consumption}
We quantify the inference energy proxy of the upper-level agentic AI, denoted by $J_{\mathrm{ag}}^{(n)}$, by accounting only for the external LLM calls involved in generating the upper-level decision within the $n$-th large timescale interval. The energy proxy is defined as
\begin{align} \label{llm energy formulation}
J_{\mathrm{ag}}^{(n)}
=
\sum_{q\in\mathcal{Q}^{(n)}}
e\!\left(m_q,\tau_{q,\mathrm{in}},\tau_{q,\mathrm{out}}\right),\ n=0,1,\cdots,N-1
\end{align}
where $\mathcal{Q}^{(n)}$ denotes the set of such external LLM calls in interval $n$, $m_q$ denotes the model/service type used by call $q$, and $\tau_{q,\mathrm{in}}$ and $\tau_{q,\mathrm{out}}$ denote the corresponding input and output token counts, respectively. $e(m,\tau_{\mathrm{in}},\tau_{\mathrm{out}})$ gives the energy proxy of one external call under model/service type $m$.

Following workload-aware LLM energy models~\cite{cavagna2026sweetspot}, a practical proxy
form is given by
\begin{equation} \label{llm energy specific}
e\!\left(m,\tau_{\mathrm{in}},\tau_{\mathrm{out}}\right)
=
\alpha_{m,0}\tau_{\mathrm{in}}
+\alpha_{m,1}\tau_{\mathrm{out}}
+\alpha_{m,2}\tau_{\mathrm{in}}\tau_{\mathrm{out}},
\end{equation}
where $\alpha_{m,0},\alpha_{m,1},\alpha_{m,2}\ge 0$ are
model-dependent proxy coefficients. This construction accounts for the fact that the energy cost of an external LLM call depends not only on the input and output token lengths, but also on their interaction.
\subsection{Constraints of Lower-level beamforming design}

The lower-level beamforming is executed per time slot $t\in\mathcal{T}$ based on the instantaneous CSI and the configuration provided by the upper level.
The CSI available at the APs can be imperfect due to estimation errors and calibration mismatches. To model CSI uncertainty, we adopt
\begin{align}
\mathbf{h}_{\ell k}^t = \widehat{\mathbf{h}}_{\ell k}^t + \Delta\mathbf{h}_{\ell k}^t,
\quad \forall \ell\in\mathcal{L}, k\in\mathcal{K}, t \in \mathcal{T},
\label{eq:csi_error_model}
\end{align}
where $\widehat{\mathbf{h}}_{\ell k}^t$ is the estimated CSI, and $\Delta\mathbf{h}_{\ell k}^t$ is the unknown CSI error. Assume the CSI error is norm-bounded:
\begin{align}
\left\|\Delta\mathbf{h}_{\ell k}^t\right\|_2 \le \delta_{\ell k}^t,
\quad \forall \ell\in\mathcal{L}, k\in\mathcal{K}, t \in \mathcal{T},
\label{eq:csi_uncertainty_set}
\end{align}
where $\delta_{\ell k}^t$ denotes the corresponding error radius. Perfect CSI is a special case of this model by setting the error radius to $\delta_{\ell k}^t=0, \forall \ell\in\mathcal{L},k\in\mathcal{K}$.

Under imperfect CSI, we define the worst-case (robust) SINR of user $k$ in time slot $t$ as
\begin{align}\label{eq:wc_sinr_def}
\mathrm{SINR}_{k}^{t,\mathrm{wc}} \!=\!\!
\inf_{ \|\Delta{\bf h}_{\ell k}^t\|\le \delta_{\ell k}^t, \forall \ell \in \mathcal{L}} & \mathrm{SINR}_k^t\!\left(\!\!\{\hat{\bf h}_{\ell k}^t \!+\!\! \Delta{\bf h}_{\ell k}^t\}_{\ell=1}^{L},\{{\bf w}_{\ell j}^t\}\!\!\right),
\nonumber \\
& \forall k\in\mathcal{K}, t\in\mathcal{T},
\end{align}
where $\hat{\bf h}_{\ell k}^{t}$ is the estimated CSI from AP $\ell$ to user $k$ in slot $t$.
Accordingly, the worst-case (robust) QoS requirement is compactly written as
\begin{align} 
\mathrm{SINR}_{k}^{t,\mathrm{wc}} \ge \Gamma_k^t,
\quad \forall k\in\mathcal{K}, t\in\mathcal{T}.
\label{eq:robust_qos_wc}
\end{align}

We also consider the per-AP transmit power constraints in each slot.
Each AP $\ell$ has a transmit power budget $P_{\ell}^{\max}$, i.e.,
\begin{align}
P_{\ell}^t \triangleq \mathbb{E}\!\left\{\left\|\mathbf{x}_{\ell}^t\right\|_2^2\right\}
\!=\! \sum_{k=1}^{K} \left\|\mathbf{w}_{\ell k}^t\right\|_2^2
\le \phi_{\ell}^t P_{\ell}^{\max},
\ \forall \ell\in\mathcal{L},~t\in\mathcal{T},
\label{eq:per_ap_power}
\end{align}
where $\phi_{\ell}^t \in [0,1]$ denotes the power-budget exposure factor of AP $\ell$ at slot $t$, which specifies the fraction of the transmit power budget $P_{\ell}^{\max}$ made available to beamforming. If $\phi_{\ell}^t = 0$, AP $\ell$ is deactivated in slot $t$; $\phi_{\ell}^t = 1$, its full transmit power budget $P_{\ell}^{\max}$ is available for beamforming.

\section{Proposed Agentic-LTPO Framework}

In this section, we apply Agentic-LTPO to solve the CF-MIMO downlink beamforming task.
Based on the nested framework in (1), we elaborate on the upper-level optimization, including the multi-agent structure and the RAG module, followed by the lower-level beamforming with a fast solver.

\subsection{Problem Formulation}

We formulate the nested optimization problem of the CF-MIMO downlink beamforming at two timescales. The pre-specified time horizon $T$ is first partitioned into $N$ large timescale intervals, each containing $T_s$ slots, i.e., $T = N T_s$. The index set of the $n$-th large interval $\mathcal{T}^{(n)}$ is defined as
\begin{equation}
\mathcal{T}^{(n)} \triangleq \{nT_s + 1, nT_s+2, \ldots, (n+1)T_s\},
\end{equation}
where $n =0,1,\cdots,N-1$.

At the beginning of the $n$-th large interval, the upper-level Agentic AI deployed at the CPU outputs a configuration vector $\boldsymbol{c}^{(n)}$ based on its overall decision strategy $\boldsymbol{\pi}$, as given by
\begin{align} \label{c definition}
\boldsymbol{c}^{(n)} \!\!=\! \boldsymbol{\pi}\left( S_{\mathrm{env}}^{(n)},P^{(n)},\mathcal{E}^{(n)} \right) \!\!=\!\! \Big( \!
\{\Gamma_k^{(n)}\}_{k\in\mathcal K},
\{\phi_{\ell}^{(n)}\}_{\ell\in\mathcal L}
\Big),
\end{align}
where $S_{\mathrm{env}}^{(n)}$ denotes the previous interval environment summary; $P^{(n)}$ represents natural language composed of the current policy profile from the operator; $\mathcal{E}^{(n)}$ denotes the cross-timescale experience maintained by the upper level; $\Gamma_k^{(n)}$ is the corresponding target level of the worst-case (robust) QoS requirement of user $k$, c.f. \eqref{eq:robust_qos_wc}; and $\phi_{\ell}^{(n)}$ is the power-budget exposure factor of AP $\ell$, c.f. \eqref{eq:per_ap_power}. As clarified in Section III-B, $\mathcal{E}^{(n)}$ provides the base historical tuples, from which the case memory used by the RAG module is constructed.


As specified in \eqref{c definition}, the decision strategy $\boldsymbol{c}^{(n)}$ remains unchanged over slots $t\in\mathcal T^{(n)}$ of the $n$-th large interval. The nested optimization problem be formulated as
\begin{subequations}\label{eq:overall_bilevel}
\begin{align}
\max_{\boldsymbol{\pi}}\quad
& \sum_{n=0}^{N-1} G\!\left(\mathrm{KPI}^{(n)}\right)
\label{eq:overall_bilevel_a}\\
\text{s.t.}\quad
& \boldsymbol{c}^{(n)}=\boldsymbol{\pi}\!\left(S_{\mathrm{env}}^{(n)},P^{(n)},\mathcal{E}^{(n)}\right),\ \forall n,
\label{eq:overall_bilevel_b}\\
& {\bf W}^{t,*}\big(\!\boldsymbol{c}^{(n)}\big)\in
\arg\min_{\{{\bf w}_{\ell k}^{t}\}}
\sum_{\ell\in\mathcal{L}}\sum_{k\in\mathcal{K}}\big\|{\bf w}_{\ell k}^{t}\big\|_2^{2},
\label{eq:overall_bilevel_c}\\
& \ \ \text{s.t.} \sum_{k\in\mathcal{K}}\!\!\big\|{\bf w}_{\ell k}^{t}\big\|_2^{2} \!\le\!  \phi_{\ell}^{(n)}\!P_{\ell}^{\max}, \forall t \!\in\! \mathcal{T}^{(n)}, \ell \!\in\! \mathcal{L},
\label{eq:overall_bilevel_d}\\
& \qquad\ \ \mathrm{SINR}_{k}^{t,\mathrm{wc}} \ge \Gamma_{k}^{(n)},\ \forall t\in\mathcal{T}^{(n)}, k\in\mathcal{K},
\label{eq:overall_bilevel_e}
\end{align}
\end{subequations}
In (\ref{eq:overall_bilevel}a), $G(\cdot)$ is a long-term performance function specified below. 
(\ref{eq:overall_bilevel}b) is the configuration vector for the lower-level optimization; see \eqref{c definition}.
(\ref{eq:overall_bilevel}c)--(\ref{eq:overall_bilevel}e) represent the lower-level problem executed at slot $t$, namely, energy-minimizing beamforming in \eqref{eq:overall_bilevel_c} under the per-AP power constraint (\ref{eq:overall_bilevel}d) and the worst-case (robust) QoS constraint (\ref{eq:overall_bilevel}e).

The upper-level objective (\ref{eq:overall_bilevel}a) is based on the performance metrics described in Section II-C, and the computation cost from the upper-level Agentic AI. With the energy proxy $J_{\mathrm{ag}}^{(n)}$ of the upper-level Agentic AI, $\mathrm{KPI}^{(n)}$ can be defined as 
\begin{align}
    & \mathrm{KPI}^{(n)}=\left( \{R^t\}_{t\in\mathcal{T}^{(n)}}, \{\mathrm{EE}^t\}_{t\in\mathcal{T}^{(n)}}, J_{\mathrm{ag}}^{(n)}  \right),\ \forall n . 
\label{eq:overall_bilevel_g}
\end{align}
where $\{R^{t},\mathrm{EE}^{t}\}_{t\in\mathcal{T}^{(n)}}$ collects the sum-rate and energy-efficiency values achieved in the $n$-th large interval $\mathcal{T}^{(n)}$.
Given $\mathrm{KPI}^{(n)}$, we design the upper-level objective as
\begin{align} \label{upper obj G}
    & G(\mathrm{KPI}^{(n)}) \!=\! \frac{\lambda_R}{T_s} \!\!\! \sum_{t \in \mathcal{T}^{(n)}} \!\!\! \frac{R^t}{R_{\mathrm{ref}}} \!+\! \frac{\lambda_{\mathrm{EE}}}{T_s} \!\!\! \sum_{t \in \mathcal{T}^{(n)}} \!\!\! \frac{\mathrm{EE}^t}{\mathrm{EE}_{\mathrm{ref}}} \!-\! \lambda_J \! \frac{J_{\mathrm{ag}}^{(n)}}{J_{\mathrm{ref}}} ,
\end{align}
where $R_{\mathrm{ref}}$, $\mathrm{EE}_{\mathrm{ref}}$, and $J_{\mathrm{ref}}$ are pre-specified positive normalization constants to rescale the different KPI components into comparable quantities.
$\lambda_R$, $\lambda_{\mathrm{EE}}$, and $\lambda_J$
are positive weighting coefficientss to balance the contributions of each KPI. 
\eqref{upper obj G} defines a long-term utility that balance communication performance and decision complexity, encouraging
policies that achieve high spectral and energy efficiency while avoiding
unnecessary upper-level reasoning overhead.



Problem \eqref{eq:overall_bilevel} captures the timescale coupling in Agentic-LTPO. 
At the large timescale, the upper level updates the configuration vector per large interval to maximize the long-term utility $G\!\big(\mathrm{KPI}^{(n)}\big)$, which balances the sum-rate, energy efficiency, and agentic AI inference cost. 
At the small timescale, the lower level allows the APs to make beamforming decisions slot-by-slot under imperfect instantaneous CSI in a distributed fashion, subject to the policy constraints specified by the upper-level configuration.

\subsection{Upper-Level Optimization with Multi-Agent Collaboration}\label{subsec:upper_level}

As shown in \eqref{eq:overall_bilevel}, the upper level determines the configuration vector $\boldsymbol{c}^{(n)}$ for the $n$-th large interval from the environment summary $S_{\mathrm{env}}^{(n)}$, the operator's policy profile $P^{(n)}$, and the accumulated experience buffer $\mathcal E^{(n)}$.\footnote{This mapping is non-trivial since the upper level must convert these heterogeneous, cross-timescale, and partially semantic inputs into a structured configuration vector, whereas the quality of the configuration can only be assessed indirectly through the lower-level responses over $t\in\mathcal T^{(n)}$ and the resulting KPI aggregation in \eqref{eq:overall_bilevel_g}.} 

\begin{figure}[!t]
	\centering
	\includegraphics[width=1\linewidth]{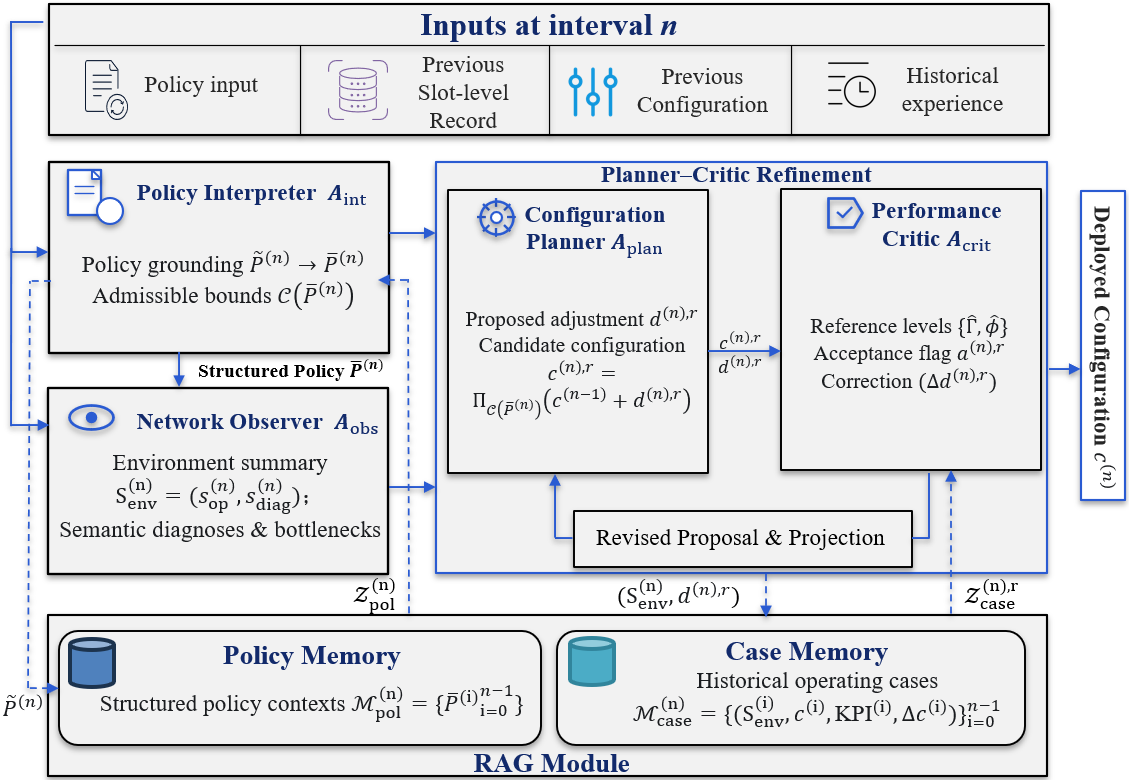}
	\caption{The upper-level multi-agent architecture of Agentic-LTPO for CF-MIMO long-term performance optimization.}
	\label{fig:agent model}
\end{figure} 

We propose a coordinated multi-agent architecture,
which leverages LLM capabilities, including long-term planning,
structured policy interpretation, and retrieval-enhanced reasoning, to process the heterogeneous inputs efficiently, as illustrated in Fig.~\ref{fig:agent model}.
To realize the mapping in \eqref{c definition} in a structured and interpretable manner, we adopt four coordinated LLM agents, namely, Policy Interpreter $A_{\mathrm{int}}$, Network Observer $A_{\mathrm{obs}}$, Performance Critic $A_{\mathrm{crit}}$, and Configuration Planner $A_{\mathrm{plan}}$, together with a RAG module.

At the beginning of the $n$-th large interval, the decision $\boldsymbol{c}^{(n)}$ is generated from the heterogeneous input $(S_{\mathrm{env}}^{(n)},P^{(n)},\mathcal{E}^{(n)})$.
To construct the information flow across intervals, we collect the per-slot records of the previous interval, as given by
\begin{align} \label{eq:trace_def_multiagent}
\mathcal{D}^{(n-1)}
= \left\{\left(\hat{\mathbf{H}}^t,\mathbf{W}^{t,*}\big(\boldsymbol{c}^{(n-1)}\big),\mathbf{u}^t\right)\right\}_{t\in\mathcal{T}^{(n-1)}},
\end{align}
where $\hat{\mathbf{H}}^t \triangleq[\hat{\mathbf{h}}_1^t, \ldots, \hat{\mathbf{h}}_K^t] \in \mathbb{C}^{LM\times K}$ is the estimated channel matrix in slot $t$ with $\hat{\mathbf{h}}_k^t \triangleq[(\hat{\mathbf{h}}_{1 k}^t)^{\intercal}, \ldots,(\hat{\mathbf{h}}_{L k}^t)^{\intercal}]^{\intercal}$, and
$\mathbf{u}^t \triangleq \big(\{\mathrm{SINR}_{k}^{t,\mathrm{wc}}\}_{k\in\mathcal K},
\{\frac{P_\ell^t}{P_\ell^{\max}}\}_{\ell\in\mathcal L}, R^t, \mathrm{EE}^t\big)$
collects the per-slot KPI measurements after deploying $\boldsymbol{c}^{(n-1)}$, with
$P_\ell^t=\sum_{k\in\mathcal K}\|{\bf w}_{\ell k}^{t}\|_2^2$.

Meanwhile, the historical experience $\mathcal{E}^{(n)}$ available to the upper level at the beginning of interval $n$ is expressed as
\begin{equation}
\mathcal{E}^{(n)}
\triangleq
\left\{
\left(
S_{\mathrm{env}}^{(i)},
\boldsymbol{c}^{(i)},
\mathrm{KPI}^{(i)}
\right)
\right\}_{i=0}^{n-1},\label{eq:experience_def_multiagent}
\end{equation}
where $\mathcal{E}^{(n)}$ serves as the cross-timescale experience used in the
upper level. For retrieval, its stored tuples are augmented in Section III-B.4)
to construct the corresponding case memory.

\subsubsection{Policy Interpreter $A_{\mathrm{int}}$}

The policy input $P^{(n)}$ from the operator is typically specified in natural language, e.g., operation guidelines, service requirements, or deployment rules, which are not directly executable by the agents since they do not provide an explicit representation of decision-related fields~\cite{manias2024intent5g}. 
We invoke the Policy Interpreter $A_{\mathrm{int}}$ to convert $P^{(n)}$ into a structured policy context in a JSON format.
$A_{\mathrm{int}}$ first produces a preliminary structured policy $\tilde{P}^{(n)}$ from $P^{(n)}$, and then refines it with the retrieved policy evidence set $\mathcal{Z}_{\mathrm{pol}}^{(n)}$ from the RAG module; see Section III-B.4).
The final structured policy context is given by
\begin{equation}
\bar{P}^{(n)}=A_{\mathrm{int}}(P^{(n)},\mathcal{Z}_{\mathrm{pol}}^{(n)}),
\label{eq:policy_interpreter}
\end{equation}
where $\bar{P}^{(n)}$ is a machine-readable policy object organized as typed key--value entries.
It encodes the policy semantics relevant to the configurations in \eqref{c definition}, including the QoS requirements $\{\Gamma_k^{(n)}\}_{k\in\mathcal K}$, the power-budget factors $\{\phi_{\ell}^{(n)}\}_{\ell\in\mathcal L}$, and the feasible decision domain.
The retrieved evidence in $\mathcal{Z}_{\mathrm{pol}}^{(n)}$ enhances the current policy input by stabilizing field extraction, resolving semantically similar policies across intervals, and improving the consistency of the admissible bounds for the robust QoS targets $\{\Gamma_k\}_{k\in\mathcal{K}}$ and the AP power-budget factors $\{\phi_\ell\}_{\ell\in\mathcal{L}}$.
Notably, $\tilde{P}^{(n)}$ is the preliminary parse of the current operator policy; $\bar{P}^{(n)}$ is the evidence-calibrated policy object used by the downstream agents. $\tilde{P}^{(n)}$ and $\bar{P}^{(n)}$ follow the same structured-policy schema.

To make the policy constraints explicit, $A_{\mathrm{int}}$ generates the feasible configuration set from the structured policy context $\bar{P}^{(n)}$, as given by
\begin{align}
&\mathcal{C}(\!\bar{P}^{(n)}\!)
\!=\!
\Big\{ \!\!
\big( \!
\{\Gamma_k\!\}_{k\in\mathcal{K}},
\{\phi_\ell\!\}_{\ell\in\mathcal{L}}
\big) \!:\! \Gamma_k \!\in\! [\Gamma_k^{\min}(\!\bar{P}^{(n)}\!),\Gamma_k^{\max}(\!\bar{P}^{(n)}\!)], \nonumber \\
& \forall k\in\mathcal{K},
\phi_\ell \!\in\! [\phi_\ell^{\min}(\bar{P}^{(n)}) , \phi_\ell^{\max}(\bar{P}^{(n)})],\ \forall \ell\in\mathcal{L}
\Big\}, 
\label{eq:CbarP_def}
\end{align}
where the bounds are determined from the JSON-structured policy context $\bar{P}^{(n)}$. 
In particular, the bounds on $\{\Gamma_k\}_{ k\in\mathcal{K}}$ specify the admissible robust QoS targets under the operator's current intent; the bounds on $\{\phi_\ell\}_{\ell \in \mathcal{L}}$ specify the admissible AP-side power-budget activation range.

Notably, $A_{\mathrm{int}}$ removes the ambiguity of $P^{(n)}$ and exposes the policy information in a standardized schema that can be accessed by the agents, i.e., $A_{\mathrm{obs}}$, $A_{\mathrm{plan}}$, or $A_{\mathrm{crit}}$.
Meanwhile, $A_{\mathrm{int}}$ links the operator's intent to the subsequent upper-level optimization, enabling $A_{\mathrm{obs}}$, $A_{\mathrm{plan}}$, $A_{\mathrm{crit}}$ and the RAG module to operate on the same structured policy $\bar{P}^{(n)}$.
If no policy evidence is retained from the RAG module,
$A_{\mathrm{int}}$ uses the retrieved policy evidence to refine $\tilde{P}^{(n)}$ and outputs $\bar{P}^{(n)}$.

\subsubsection{Network Observer $A_{\mathrm{obs}}$}

The role of $A_{\mathrm{obs}}$ is to translate the performance of the previous large interval to a compact state representation usable by the current upper-level optimization.
Based on $\mathcal{D}^{(n-1)}$, $\boldsymbol{c}^{(n-1)}$ and $\bar{P}^{(n)}$ from $A_{\mathrm{int}}$, the network observer $A_{\mathrm{obs}}$ produces the environment summary on the basis of large intervals, as given by
\begin{equation}
S_{\mathrm{env}}^{(n)}
=
A_{\mathrm{obs}}\!\left(\mathcal{D}^{(n-1)},\boldsymbol{c}^{(n-1)},\bar{P}^{(n)}\right)
=
\Big(
\mathbf{s}_{\mathrm{op}}^{(n)},
\mathbf{s}_{\mathrm{diag}}^{(n)}
\Big),
\label{eq:observer_state_structure}
\end{equation}
where $\mathbf{s}_{\mathrm{op}}^{(n)}\triangleq\left(\{\Gamma_{k,\mathrm{obs}}^{(n)}\}_{k\in\mathcal K},\{\phi_{\ell,\mathrm{obs}}^{(n)}\}_{\ell\in\mathcal L}\right)$ with
$\Gamma_{k,\mathrm{obs}}^{(n)}=\min_{t\in\mathcal{T}^{(n-1)}} \mathrm{SINR}_{k}^{t,\mathrm{wc}}$, \(\forall k\in\mathcal K\) and $\phi_{\ell,\mathrm{obs}}^{(n)}=\max_{t\in\mathcal{T}^{(n-1)}} \frac{P_\ell^t}{P_\ell^{\max}}$, \(\forall \ell\in\mathcal L\). Particularly, $\Gamma_{k,\mathrm{obs}}^{(n)}$ and $\phi_{\ell,\mathrm{obs}}^{(n)}$ are user $k$'s worst-case (robust) QoS level and AP $\ell$'s largest normalized power usage, respectively.
$\mathbf{s}_{\mathrm{diag}}^{(n)}$ is a concise semantic diagnosis generated from the same inputs as $\mathbf{s}_{\mathrm{op}}^{(n)}$, which summarizes the causes behind the observed operating bottlenecks and the resulting KPI profile, e.g., which users repeatedly approach their QoS limits and which APs frequently operate close to their power budgets.

Notably, the environment summary $S_{\mathrm{env}}^{(n)}$ in \eqref{eq:observer_state_structure} is not intended to restate the operator's previous objective. Instead, it summarizes the current operating state reached by the system after executing $\boldsymbol{c}^{(n-1)}$, so that $A_{\mathrm{plan}}$ and $A_{\mathrm{crit}}$ can determine how to move from the current state toward the new policy intent encoded in $\bar{P}^{(n)}$.
$A_{\mathrm{obs}}$ links lower-level solutions to upper-level adaptation: It compresses the per-slot performance $\{\mathbf{u}^t\}_{t\in\mathcal{T}^{(n-1)}}$ contained in $\mathcal{D}^{(n-1)}$ into a structured state $S_{\mathrm{env}}^{(n)}=(\mathbf{s}_{\mathrm{op}}^{(n)},\mathbf{s}_{\mathrm{diag}}^{(n)})$, exposes the bottlenecks that limit the current setting, and provides the state input $S_{\mathrm{env}}^{(n)}$ for the other agents and the RAG module.

\subsubsection{Configuration Planner $A_{\mathrm{plan}}$ and Performance Critic $A_{\mathrm{crit}}$}

After obtaining $\bar{P}^{(n)}$ and $S_{\mathrm{env}}^{(n)}$ from $A_{\mathrm{int}}$ and $A_{\mathrm{obs}}$, respectively, the upper level determines the configuration for interval $n$ through an effective planner--critic refinement procedure.
$A_{\mathrm{plan}}$ and $A_{\mathrm{crit}}$ perform up to $R$ refinement rounds within every large interval to obtain $\boldsymbol{c}^{(n)}$ for the lower level.
The key is to separate the candidate configuration $\boldsymbol{c}^{(n),r}$ from evidence-based verification, where $ r = 0,1,\cdots,R-1$ denotes the adjustment at the $r$-th refinement round within interval $n$. 
Specifically, $A_{\mathrm{plan}}$ proposes a structured adjustment $\boldsymbol{d}^{(n),r}$ using the current policy context and operating state; $A_{\mathrm{crit}}$ evaluates the resulting candidate $\boldsymbol{c}^{(n),r}$ using the case evidence $\mathcal{Z}_{\mathrm{case}}^{(n),r}$ retrieved by the RAG module, to determine whether  $\boldsymbol{c}^{(n),r}$ is sufficiently supported by historically similar operating conditions before deployment in the lower level.
This design provides a correction mechanism before the configuration is sent to the lower-level fast solver.

At the beginning of interval $n$, with the previous configuration vector $\boldsymbol{c}^{(n-1)}$, the environment summary $S_{\mathrm{env}}^{(n)}$, and the structured policy context $\bar{P}^{(n)}$, the configuration planner $A_{\mathrm{plan}}$ first proposes the initial adjustment vector, as given by
\begin{equation}
\boldsymbol{d}^{(n),0}
=
A_{\mathrm{plan}}
\!\left(
S_{\mathrm{env}}^{(n)},
\bar{P}^{(n)},
\boldsymbol{c}^{(n-1)}
\right),
\label{eq:planner_init}
\end{equation}
where $\boldsymbol{d}^{(n),0}$ is the planner-generated adjustment used in the first refinement round. For notational simplicity, every planner-generated adjustment $\boldsymbol{d}^{(n),r}$, including the $r=0$, is written componentwise as
$\boldsymbol{d}^{(n),r}
=
\big(
\{\Delta \Gamma_k^{(n),r}\}_{k\in\mathcal{K}},
\{\Delta \phi_\ell^{(n),r}\}_{\ell\in\mathcal{L}}
\big)$,
$r=0,1,\ldots,R-1$, where $\Delta \Gamma_k^{(n),r}$ is the generated adjustment of the robust QoS target of user $k$ and $\Delta \phi_\ell^{(n),r}$ is the generated adjustment of the power-budget exposure factor of AP $\ell$.

At refinement round $r$, $A_{\mathrm{plan}}$ forms the candidate configuration:
\begin{equation}
\boldsymbol{c}^{(n),r}
=
\Pi_{\mathcal{C}(\bar{P}^{(n)})}
\!\left(
\boldsymbol{c}^{(n-1)} + \boldsymbol{d}^{(n),r}
\right),\ r=0,1,\cdots,R-1,
\label{eq:planner_candidate}
\end{equation}
where $\Pi_{\mathcal{C}(\bar{P})}(\cdot)$ is the Euclidean projection onto the set in \eqref{eq:CbarP_def}, and implemented by the clipping operation:
$\forall k\in\mathcal{K}, \ell\in\mathcal{L}$,
\begin{align}
\Gamma_k^{(n),r}
 \!\!=\!\!
\min\!\Big\{ \!
\Gamma_k^{\max}(\!\bar{P}^{(n)}\!),
\max\!\big\{ \!
\Gamma_k^{\min}(\!\bar{P}^{(n)}\!),
\Gamma_k^{(n\!-\!1)} \!\!+\! \Delta\Gamma_k^{(n),r} \!
\big\} \!\!
\Big\};
\nonumber
\\
\phi_{\ell}^{(n),r}
 \!\!=\!\!
\min\!\Big\{ \!
\phi_{\ell}^{\max}(\!\bar{P}^{(n)}\!),
\max\!\big\{ \!
\phi_{\ell}^{\min}(\!\bar{P}^{(n)}\!),
\phi_{\ell}^{(n\!-\!1)} \!\!+\! \Delta\phi_{\ell}^{(n),r} \!
\big\} \!\!
\Big\}.
\label{eq:phi_clip}
\end{align}
The clipping is required because $\boldsymbol{d}^{(n),r}$ cannot guarantee policy feasibility. By projecting the raw candidate $\boldsymbol{c}^{(n-1)}+\boldsymbol{d}^{(n),r}$ onto $\mathcal{C}(\bar{P}^{(n)})$, we ensure that the candidate configuration $\boldsymbol{c}^{(n),r}$ transmitted to $A_{\mathrm{crit}}$ satisfies the policy bounds in \eqref{eq:CbarP_def}.

At refinement round $r$ of interval $n$, the critic $A_{\mathrm{crit}}$ evaluates $\boldsymbol{c}^{(n),r}$ using the case evidence set $\mathcal{Z}_{\mathrm{case}}^{(n),r}$ retrieved by the RAG module, as will be described in Section III-B.4). Based on the normalized weights associated with the retrieved cases, $A_{\mathrm{crit}}$ forms the empirical reference levels $\{\widehat{\Gamma}_k^{(n),r}\}_{k\in\mathcal{K}}$ and
$\{\widehat{\phi}_\ell^{(n),r}\}_{\ell\in\mathcal{L}}$ from the historical support statistics $\{\Gamma_{k,\mathrm{obs}}^{(j)}\}$ and $\{\phi_{\ell,\mathrm{obs}}^{(j)}\}$ contained in the retained environment
summaries $\{S_{\mathrm{env}}^{(j)}\}_{j\in\mathcal{I}_{\mathrm{case}}^{(n),r}}$; see Section III-B.4). Then, $A_{\text{crit}}$ determines
\begin{equation}
\left(\!
a^{(n),r},
\Delta \boldsymbol{d}^{(n),r}
\!\right)
\!\!=\!\!
A_{\mathrm{crit}}
\!\!\left(\!
S_{\mathrm{env}}^{(n)},
\bar{P}^{(n)},
\boldsymbol{c}^{(n),r},
\boldsymbol{d}^{(n),r},
\mathcal{Z}_{\mathrm{case}}^{(n),r}
\!\right),
\label{eq:critic_output_new}
\end{equation}
where $a^{(n),r}\in\{0,1\}$ denotes an acceptance flag; and $\Delta \boldsymbol{d}^{(n),r}$ denotes the correction generated by $A_{\mathrm{crit}}$, which is used by $A_{\mathrm{plan}}$ to revise $\boldsymbol{d}^{(n),r}$ in the next refinement round $r+1$ of the $n$-th large interval.
Since the candidate $\boldsymbol{c}^{(n),r}$ has already been projected
onto $\mathcal{C}(\bar{P}^{(n)})$, $A_{\mathrm{crit}}$ checks whether it remains within the empirically supported neighborhood of the retrieved reference levels, i.e.,
$\forall k \in \mathcal{K}, \ell \in \mathcal{L}$,
\begin{align}\label{eq:critic_accept}
 a^{(n),r}=1,\ \text{if}\ &\left|\Gamma_k^{(n),r}-\widehat{\Gamma}_k^{(n),r}\right|
\le \varepsilon_{\Gamma,k}^{(n),r},\ \text{and} \nonumber \\
 &\left|\phi_\ell^{(n),r}-\widehat{\phi}_\ell^{(n),r}\right|
\le \varepsilon_{\phi,\ell}^{(n),r}
\end{align}
where $\varepsilon_{\Gamma,k}^{(n),r}\ge 0$ and
$\varepsilon_{\phi,\ell}^{(n),r}\ge 0$ are component-wise empirically supported tolerances around the retrieved reference levels. 
Specifically, $\varepsilon_{\Gamma,k}^{(n),r}$ bounds the admissible deviation between the candidate robust QoS target $\Gamma_k^{(n),r}$ and its retrieved reference $\widehat{\Gamma}_k^{(n),r}$, while $\varepsilon_{\phi,\ell}^{(n),r}$ bounds the admissible deviation between the candidate AP power-budget exposure factor $\phi_\ell^{(n),r}$ and its retrieved reference $\widehat{\phi}_\ell^{(n),r}$.
Otherwise, $A_{\mathrm{crit}}$ returns a correction $\Delta\boldsymbol{d}^{(n),r}$ that pushes the candidate $\boldsymbol{c}^{(n),r}$ back toward the empirically supported region:
\begin{align}\label{eq:delta_d_def}
&\Delta \boldsymbol{d}^{(n),r}
=\\
&\Big(
\{\rho_\Gamma(\widehat{\Gamma}_k^{(n),r}-\Gamma_k^{(n),r})\}_{k\in\mathcal K},
\{\rho_\phi(\widehat{\phi}_\ell^{(n),r}-\phi_\ell^{(n),r})\}_{\ell\in\mathcal L}
\Big), \nonumber
\end{align} 
where $\rho_\Gamma,\rho_\phi>0$ are correction coefficients.

If $a^{(n),r}=1$, the candidate $\boldsymbol{c}^{(n),r}$ is accepted and deployed as the configuration vector of interval $n$. Otherwise, $A_{\mathrm{plan}}$ revises its adjustment $\boldsymbol{d}^{(n),r}$ according to $A_{\mathrm{crit}}$'s feedback:
\begin{equation}
\boldsymbol{d}^{(n),r+1}
=
A_{\mathrm{plan}}
\!\left(
S_{\mathrm{env}}^{(n)},
\bar{P}^{(n)},
\boldsymbol{c}^{(n-1)},
\boldsymbol{d}^{(n),r},
\Delta \boldsymbol{d}^{(n),r}
\right),
\label{eq:planner_revise}
\end{equation}
Then, the same verification process as in \eqref{eq:planner_candidate}--\eqref{eq:critic_accept} is repeated.
Let $r^\star$ be the first refinement round achieving $a^{(n),r^\star}=1$. The deployed configuration is given by
\begin{equation}\label{eq:planner_accept}
\boldsymbol{c}^{(n)} = \boldsymbol{c}^{(n),r^\star}.
\end{equation}
If no adjustment is accepted within $R$ refinement rounds, $A_{\mathrm{plan}}$ deploys $\boldsymbol{c}^{(n),R-1}$ as the output. 
Thus, the upper level does not directly map $(S_{\mathrm{env}}^{(n)},\bar{P}^{(n)})$ to a configuration in one step; instead, it determines $\boldsymbol{c}^{(n)}$ through a planner–critic refinement process, which is robust to decision errors and aligned with the policy-aware deployment requirement of Agentic-LTPO.

\subsubsection{RAG module}\label{subsubsec:rag_system}

To support the planner--critic refinement in Section III-B.3), we equip the upper-level critic agent $A_{\mathrm{crit}}$ with the RAG module to provide policy- and experience-based evidence throughout the refinement rounds.
The RAG module enables $A_{\mathrm{int}}$ and $A_{\mathrm{plan}}$ to reuse previous policy contexts and historical operating cases under similar operating regimes, making the update from $\boldsymbol{c}^{(n-1)}$ to $\boldsymbol{c}^{(n)}$ stable and interpretable.

The RAG module maintains two memories, namely, a policy memory and a case memory.
The policy memory stores the structured policy contexts generated by $A_{\mathrm{int}}$, as given by
\begin{equation}
\mathcal{M}_{\mathrm{pol}}^{(n)}
\triangleq
\left\{
\bar{P}^{(i)}
\right\}_{i=0}^{n-1}.
\label{eq:rag_policy_memory}
\end{equation}
The case memory is constructed by augmenting the historical tuples stored in the cross-timescale experience buffer $\mathcal{E}^{(n)}$ with the deployed configuration changes, as given by
\begin{equation}
\mathcal{M}_{\mathrm{case}}^{(n)}
\triangleq
\left\{
\Big(
S_{\mathrm{env}}^{(i)},
\boldsymbol{c}^{(i)},
\mathrm{KPI}^{(i)},
\Delta\boldsymbol{c}^{(i)}
\Big)
\right\}_{i=0}^{n-1},
\label{eq:rag_case_memory}
\end{equation}
where $\Delta\boldsymbol{c}^{(i)} = \boldsymbol{c}^{(i)}-\boldsymbol{c}^{(i-1)}, i\ge 1 $ with $\Delta\boldsymbol{c}^{(0)}$ initialized an all-zero vector, and $\Delta\boldsymbol{c}^{(i)}$ records the deployed configuration change of interval $i$ with $i = 0,1,\cdots,n-1$.

As designed in Sections III-B.1) to III-B.3), the policy memory $\mathcal{M}_{\mathrm{pol}}^{(n)}$ stores the historical structured policy contexts produced by $A_{\mathrm{int}}$. The case memory $\mathcal{M}_{\mathrm{case}}^{(n)}$ is updated after each interval by recording the observer summary $S_{\mathrm{env}}^{(i)}$, together with the deployed configuration $\boldsymbol{c}^{(i)}$, realized KPI tuple $\mathrm{KPI}^{(i)}$, and resulting configuration change $\Delta\boldsymbol{c}^{(i)}$. 

The RAG module is used online by $A_{\mathrm{crit}}$ in \eqref{eq:critic_output_new}.
As described in Section III-B.1), $A_{\mathrm{int}}$ first forms a preliminary structured policy $\tilde{P}^{(n)}$ and then retrieves a raw policy evidence set $\widetilde{\mathcal{Z}}_{\mathrm{pol}}^{(n)}$ from $\mathcal{M}_{\mathrm{pol}}^{(n)}$:
\begin{equation}
\widetilde{\mathcal{Z}}_{\mathrm{pol}}^{(n)}
=
\operatorname{TopK}_{K_{\mathrm{pol}}}
\left\{
\bar{P}^{(i)}\in\mathcal{M}_{\mathrm{pol}}^{(n)}: \mathrm{sim}\!\left(\tilde{P}^{(n)},\bar{P}^{(i)}\right)
\right\},
\label{eq:rag_retrieve_policy_raw}
\end{equation}
where $\mathrm{sim}(\cdot,\cdot)$ gives the cosine similarity between the embeddings of the normalized records converted from $\tilde{P}^{(n)}$ and $\bar{P}^{(i)}$, produced by an embedding API\footnote{In practice, before embedding, each compared object is converted into a
typed normalized record under a unified schema.}; $K_{\mathrm{pol}}$ is the number of recalled policy items before filtering; $\operatorname{TopK}_{K}\!\left\{x_i:s_i\right\}$ returns the $K$ items $x_i$ with the highest scores $s_i$.

The retrieved policy evidence set is then determined as
\begin{equation}
\mathcal{Z}_{\mathrm{pol}}^{(n)}
=
\left\{
\bar{P}^{(i)}\in\widetilde{\mathcal{Z}}_{\mathrm{pol}}^{(n)}
:
\mathrm{sim}\!\left(\tilde{P}^{(n)},\bar{P}^{(i)}\right) \ge \tau_{\mathrm{pol}}
\right\},
\label{eq:rag_retrieve_policy}
\end{equation}
where $\tau_{\mathrm{pol}}$ is the predesigned threshold.
This policy retrieval is implemented with a nearest-neighbor search over the stored structured policy contexts in $\mathcal{M}_{\mathrm{pol}}^{(n)}$, followed by threshold-based filtering at $\tau_{\mathrm{pol}}$.
The evidence in $\mathcal{Z}_{\mathrm{pol}}^{(n)}$ is supplied to $A_{\mathrm{int}}$ to revise the preliminary fields of $\tilde{P}^{(n)}$ and produce the structured policy context $\bar{P}^{(n)}$ in \eqref{eq:policy_interpreter}.

At refinement round $r$ of the $n$-th large timescale interval, $A_{\mathrm{crit}}$ retrieves a case evidence set from $\mathcal{M}_{\mathrm{case}}^{(n)}$ to perform corrections.
To jointly evaluate the elements in \eqref{eq:rag_case_memory}, we define the following hybrid retrieval score:
\begin{align}
\eta_{\mathrm{case}}^{(i),r} 
&= 
\lambda_{\mathrm{op}} \mathrm{sim} \left( \mathbf{s}_{\mathrm{op}}^{(n)}, \mathbf{s}_{\mathrm{op}}^{(i)} \right) + 
\lambda_{\mathrm{diag}} \mathrm{sim} \left( \mathbf{s}_{\mathrm{diag}}^{(n)}, \mathbf{s}_{\mathrm{diag}}^{(i)} \right) \nonumber\\
&+ 
\lambda_{\mathrm{dir}} \mathrm{sim} \left( \boldsymbol{d}^{(n),r}, \Delta \boldsymbol{c}^{(i)} \right),
\label{eq:rag_case_score}
\end{align}
where $\lambda_{\mathrm{op}},\lambda_{\mathrm{diag}},\lambda_{\mathrm{dir}}\ge 0$ and $\lambda_{\mathrm{op}}+\lambda_{\mathrm{diag}}+\lambda_{\mathrm{dir}}=1$.
If $\|\Delta\boldsymbol{c}^{(i)}\|_2=0$, then $\mathrm{sim} ( \boldsymbol{d}^{(n),r}, \Delta \boldsymbol{c}^{(i)} ) =0$.
Based on $\eta_{\mathrm{case}}^{(i),r}$, the raw case retrieval $\widetilde{\mathcal{Z}}_{\mathrm{case}}^{(n),r}$ is defined as 
\begin{equation}
\widetilde{\mathcal{Z}}_{\mathrm{case}}^{(n),r}
\!\!=\!\!
\operatorname{TopK}_{K_{\mathrm{case}}} \!\!
\left\{ \!\!
\Big( \!\!
S_{\mathrm{env}}^{(i)},
\boldsymbol{c}^{(i)},
\mathrm{KPI}^{(i)},
\Delta\boldsymbol{c}^{(i)}
\!\! \Big)
\!\!\in\!\!
\mathcal{M}_{\mathrm{case}}^{(n)}
\!:\!
\eta_{\mathrm{case}}^{(i),r} \!
\right\},
\label{eq:rag_retrieve_case_raw}
\end{equation}
where $K_{\mathrm{case}}$ denotes the number of recalled case items before filtering. With threshold-based filtering, the retrieved case evidence set is defined as 
\begin{align}
\mathcal{Z}_{\mathrm{case}}^{(n),r}
&=
\Big\{\, 
\Big(
S_{\mathrm{env}}^{(i)},
\boldsymbol{c}^{(i)},
\mathrm{KPI}^{(i)},
\Delta\boldsymbol{c}^{(i)}
\Big)
\in
\widetilde{\mathcal{Z}}_{\mathrm{case}}^{(n),r}
:\;
\nonumber\\
&
\boldsymbol{c}^{(i)}\in\mathcal{C}\!\Big(\bar{P}^{(n)}\Big),\mathrm{sim} \left( \boldsymbol{d}^{(n),r}, \Delta \boldsymbol{c}^{(i)} \right) \ge \tau_{\mathrm{dir}}
\Big\},
\label{eq:rag_retrieve_case}
\end{align}
where $\tau_{\mathrm{dir}}$ is the threshold for directional consistency.
The case retrieval is implemented as a hybrid search in $\mathcal{M}_{\mathrm{case}}^{(n)}$: a top-$K$ recall is performed according to \eqref{eq:rag_case_score} and refined by feasibility and directional consistency filtering under the policy-constrained domain $\mathcal{C}\!\left(\bar{P}^{(n)}\right)$.

The retrieved evidence in \eqref{eq:rag_retrieve_case} is used by $A_{\mathrm{crit}}$ to form the empirical reference levels in \eqref{eq:critic_accept}. Let $\mathcal{I}_{\mathrm{case}}^{(n),r}=\{j:\;(S_{\mathrm{env}}^{(j)},\boldsymbol{c}^{(j)},\mathrm{KPI}^{(j)},\Delta\boldsymbol{c}^{(j)})\in\mathcal{Z}_{\mathrm{case}}^{(n),r}\}$. We define the normalized weight as
\begin{equation}
\omega_j^{(n),r}
=
\frac{
\exp\!\big(\eta_{\mathrm{case}}^{(j),r}\big)
}{
\sum_{i\in\mathcal{I}_{\mathrm{case}}^{(n),r}}
\exp\!\big(\eta_{\mathrm{case}}^{(i),r}\big)
},
\label{eq:rag_case_weight}
\end{equation}
With $\omega_j^{(n),r},\forall j \in \mathcal{I}_{\mathrm{case}}^{(n),r}$, the reference levels in \eqref{eq:critic_accept} are constructed as
\begin{align} \label{widehat Gamma}
\widehat{\Gamma}_{k}^{(n),r}
=
\sum_{j\in\mathcal{I}_{\mathrm{case}}^{(n),r}}
\omega_j^{(n),r}\,
\Gamma_{k,\mathrm{obs}}^{(j)}, 
\quad \forall k\in\mathcal{K};
\end{align}
\begin{align} \label{widehat phi}
\widehat{\phi}_{\ell}^{(n),r}
=
\sum_{j\in\mathcal{I}_{\mathrm{case}}^{(n),r}}
\omega_j^{(n),r}\,
\phi_{\ell,\mathrm{obs}}^{(j)},
\quad \forall \ell\in\mathcal{L}.
\end{align}
where $\Gamma_{k,\mathrm{obs}}^{(j)}$ and $\phi_{\ell,\mathrm{obs}}^{(j)}$ are user $k$'s worst-case (robust) QoS level and AP $\ell$'s largest normalized power usage in $S_{\mathrm{env}}^{(j)}$, respectively.
$A_{\mathrm{crit}}$ does not need to search the entire experience set $\mathcal{E}^{(n)}$; it grounds its acceptance decision with evidence from historical similar cases and deployed adjustments.
If the retrieved case evidence set is empty ($\mathcal{Z}_{\mathrm{case}}^{(n),r}=\emptyset$), $A_{\mathrm{crit}}$ falls back
to $\mathbf{s}_{\mathrm{op}}^{(n)}$ and sets
\[
\widehat{\Gamma}_{k}^{(n),r}
=
\Gamma_{k,\mathrm{obs}}^{(n)},\quad \forall k\in\mathcal K;
\qquad
\widehat{\phi}_{\ell}^{(n),r}
=
\phi_{\ell,\mathrm{obs}}^{(n)},\quad \forall \ell\in\mathcal L.
\]
In this case, the weighting step in \eqref{eq:rag_case_weight} is skipped.

\subsection{Lower-Level Optimization}

At each time slot $t$, for the lower level, we design the downlink beamformers $\{{\bf w}_{\ell k}^t\}$ to minimize the total transmit energy. 
Based on the configuration $\boldsymbol{c}^{(n)}$, the lower level adjusts the target robust QoS levels $\{\Gamma_k^{(n)}\}_{k\in\mathcal K}$ and the AP power-budget exposure factors $\{\phi_\ell^{(n)}\}_{\ell\in\mathcal L}$ in the slot-level constraints in slot $t\in\mathcal T^{(n)}$.
The energy minimization problem for the lower level at slot $t$ is formulated as
\begin{subequations}\label{eq:slot_energy_min}
\begin{align}
\min_{\{{\bf w}_{\ell k}^{t}\}}~~
& \sum_{\ell\in\mathcal{L}}\sum_{k\in\mathcal{K}}\big\|{\bf w}_{\ell k}^{t}\big\|_2^2
\label{eq:slot_energy_min_obj}\\
\mathrm{s.t.}~~
& \sum_{k\in\mathcal{K}}\big\|{\bf w}_{\ell k}^{t}\big\|_2^2 \le \phi_{\ell}^{(n)}\!P_{\ell}^{\max},
\quad \forall \ell\in\mathcal{L},
\label{eq:slot_power_con}\\
& \mathrm{SINR}_{k}^{t,\mathrm{wc}} \ge \Gamma_{k}^{(n)}, \quad \forall k\in\mathcal{K},
\label{eq:slot_qos_switch_k}
\end{align}
\end{subequations}
To solve \eqref{eq:slot_energy_min}, we adopt a zero-forcing (ZF) beamforming structure (even under imperfect CSI) to obtain a computationally efficient design for the lower level.

Let ${\bf h}_k^t \triangleq \big[({\bf h}_{1k}^t)^{\intercal},\ldots,({\bf h}_{Lk}^t)^{\intercal}\big]^{\intercal} \in \mathbb{C}^{LM}$
and
$\hat{\bf h}_k^t \triangleq \big[(\hat{\bf h}_{1k}^t)^{\intercal},\ldots,(\hat{\bf h}_{Lk}^t)^{\intercal}\big]^{\intercal}$. 
The unnormalized ZF directions are constructed via the right pseudo-inverse of $\hat{\bf H}^t$, as given by
\begin{equation}
\tilde{\bf V}^t \triangleq \hat{\bf H}^t\big((\hat{\bf H}^{t})^H\hat{\bf H}^t\big)^{-1}
= \big[\tilde{\bf v}_1^t,\ldots,\tilde{\bf v}_K^t\big],
\label{eq:zf_pinv_main}
\end{equation}
which yields the ZF property over the estimated channels, i.e., 
\begin{equation}
(\hat{\bf h}_k^t)^H \tilde{\bf v}_j^t = 0,\quad \forall k\in\mathcal{K},~\forall j\in\mathcal{K}\backslash\{k\}.
\label{eq:zf_property_main}
\end{equation}
We normalize each ZF direction as ${\bf v}_k^t \triangleq \tilde{\bf v}_k^t/\|\tilde{\bf v}_k^t\|_2$ and parameterize the beamformer as
\begin{equation}
{\bf w}_k^t = \sqrt{p_k^t}\,{\bf v}_k^t,\quad p_k^t\ge 0,\ \forall k\in\mathcal{K}.
\label{eq:zf_param_main}
\end{equation}
Let ${\bf v}_k^t=[({\bf v}_{1k}^t)^{\intercal},\ldots,({\bf v}_{Lk}^t)^{\intercal}]^{\intercal}$ denote the AP-wise partition corresponding to $L$ blocks of size $M$. Then, the per-AP power in time slot $t$ becomes
\begin{equation}
\sum_{k\in\mathcal{K}}\|{\bf w}_{\ell k}^t\|_2^2
=\sum_{k\in\mathcal{K}} p_k^t \|{\bf v}_{\ell k}^t\|_2^2,\quad \forall \ell\in\mathcal{L}.
\label{eq:zf_per_ap_power_main}
\end{equation}
The beamforming design in \eqref{eq:slot_energy_min} reduces to optimizing the power allocation vector ${\bf p}^t \triangleq [p_1^t,\ldots,p_K^t]^{\intercal}$.

To handle the imperfect CSI, we take the worst-case (robust) SINR into consideration. For user $k$, we define
\begin{equation}
a_k^t \triangleq \left|\sum_{\ell\in\mathcal{L}} (\hat{\bf h}_{\ell k}^t)^H {\bf v}_{\ell k}^t\right|,
\quad
\zeta_{k,j}^t \triangleq \sum_{\ell\in\mathcal{L}} \delta_{\ell k}^t \|{\bf v}_{\ell j}^t\|_2.
\label{eq:def_a_zeta}
\end{equation}
Based on \eqref{eq:def_a_zeta}, applying the triangle inequality and the Cauchy-Schwarz inequality, we obtain the following worst-case bounds of
$|\sum_{\ell\in\mathcal{L}}({\bf h}_{\ell k}^t)^H{\bf v}_{\ell k}^t|$
and
$|\sum_{\ell\in\mathcal{L}}({\bf h}_{\ell k}^t)^H{\bf v}_{\ell j}^t|$:
\begin{subequations}\label{eq:wc_bounds_zf}
\begin{align}
\min_{\{\Delta{\bf h}_{\ell k}^t: \|\Delta{\bf h}_{\ell k}^t\|_2\le \delta_{\ell k}^t\}}
\left|\sum_{\ell\in\mathcal{L}} ({\bf h}_{\ell k}^t)^H {\bf v}_{\ell k}^t\right|
\ge \big(a_k^t-\zeta_{k,k}^t\big)_+;
\label{eq:wc_sig_lb}\\
\max_{\{\Delta{\bf h}_{\ell k}^t: \|\Delta{\bf h}_{\ell k}^t\|_2\le \delta_{\ell k}^t\}}\!\!
\left|\sum_{\ell\in\mathcal{L}} ({\bf h}_{\ell k}^t)^H {\bf v}_{\ell j}^t\!\right|
\!\le\! \zeta_{k,j}^t, \forall j \!\in\! \mathcal{K}\backslash\{\!k\!\},
\label{eq:wc_int_ub}
\end{align}
\end{subequations}
where $(x)_+ \triangleq \max\{x,0\}$. Accordingly, under the strict global ZF structure in \eqref{eq:zf_param_main} (i.e., ${\bf w}_k^t=\sqrt{p_k^t}{\bf v}_k^t$), a closed-form lower bound for the worst-case (robust) SINR in \eqref{eq:wc_sinr_def} is given by
\begin{equation}\label{eq:wc_sinr_lb}
\gamma_k^t
\!\triangleq\!
\frac{p_k^t\big(a_k^t-\zeta_{k,k}^t\big)_+^2}{
(\sigma_k^t)^2 \!+\! \sum_{j\in\mathcal{K}\backslash\{k\}} p_j^t\big(\zeta_{k,j}^t\big)^2},\ {\bf p}^t \!\triangleq\! [p_1^t,\ldots,p_K^t]^T,
\end{equation}
which guarantees that $\mathrm{SINR}_{k}^{t,\mathrm{wc}} \ge \gamma_k^t$.

With \eqref{eq:wc_sinr_lb}, Problem \eqref{eq:slot_energy_min}
can be equivalently cast as
\begin{subequations}\label{eq:slot_energy_min_zf_wc}
\begin{align}
\min_{{\bf p}^t\succeq{\bf 0}}\quad
& \sum_{k\in\mathcal{K}} p_k^t
\label{eq:slot_energy_min_zf_wc_a}\\
\mathrm{s.t.}\quad
& \sum_{k\in\mathcal{K}} p_k^t\big\|{\bf v}_{\ell k}^t\big\|_2^2 \le \phi_{\ell}^{(n)}\!P_{\ell}^{\max},
\quad \forall \ell\in\mathcal{L},
\label{eq:slot_energy_min_zf_wc_b}\\
& \gamma_k^t \ge \Gamma_k^{(n)},
\quad \forall k\in\mathcal{K},
\label{eq:slot_energy_min_zf_wc_c}
\end{align}
\end{subequations}
which is a linear programming problem since the fractional constraint (\ref{eq:slot_energy_min_zf_wc}c) can be transformed into an affine inequality with respect to ${\bf p}^t$. Hence, \eqref{eq:slot_energy_min_zf_wc} can be efficiently solved using convex optimization solvers, e.g., CVX~\cite{cvx}. With the optimal solution of  \eqref{eq:slot_energy_min_zf_wc} and the ZF beamforming structure, 
this approach is efficient and well-suited for the latency-sensitive lower-level optimization at each slot.

\begin{algorithm}[htbp]
    \caption{Implementation of Agentic-LTPO}
    \label{alg:agentic_ltpo}
    \begin{algorithmic}[1]
    \STATE \textbf{Input:} $T$, $N$, $T_s$, $R$, $\{P^{(n)}\}_{n=0}^{N-1}$, $\{\gamma^{(n)}\}_{n=0}^{N-1}$, $\lambda_R$, $\lambda_{\mathrm{EE}}$, $\lambda_J$, $R_{\mathrm{ref}}$, $\mathrm{EE}_{\mathrm{ref}}$, $J_{\mathrm{ref}}$, $K_{\text{pol}}$, $\tau_{\text{pol}}$, $K_{\mathrm{case}}$, $\tau_{\mathrm{dir}}$, $\lambda_{\mathrm{op}}$, $\lambda_{\mathrm{diag}}$, $\lambda_{\mathrm{dir}}$, $\mathcal D^{(-1)}$, $\mathcal{M}_{\mathrm{pol}}^{(0)}$, $\mathcal{M}_{\mathrm{case}}^{(0)}$, $\mathcal E^{(0)}$, $\boldsymbol c^{(-1)}$
    \FOR{$n=0,1,\ldots,N-1$}
        \STATE $A_{\mathrm{int}}$ forms $\tilde{P}^{(n)}$ from $P^{(n)}$, followed by the RAG module generates policy evidence with \eqref{eq:rag_retrieve_policy_raw}--\eqref{eq:rag_retrieve_policy}. $A_{\mathrm{int}}$ generates $\bar P^{(n)}$ and $\mathcal C(\bar P^{(n)})$ with \eqref{eq:policy_interpreter} and \eqref{eq:CbarP_def}. The RAG module then updates the policy memory with $\bar P^{(n)}$
        \STATE $A_{\mathrm{obs}}$ summarizes the environment summary $S_{\mathrm{env}}^{(n)}$ from $\mathcal D^{(n-1)}$, $\boldsymbol c^{(n-1)}$, and $\bar P^{(n)}$ with \eqref{eq:observer_state_structure}
        \STATE $A_{\mathrm{plan}}$ initializes the adjustment $\boldsymbol d^{(n),0}$ by \eqref{eq:planner_init}
        \FOR{$r=0,1,\ldots,R-1$}
            \STATE $A_{\mathrm{plan}}$ forms the candidate $\boldsymbol c^{(n),r}$ by \eqref{eq:planner_candidate}
            \STATE The RAG generates case evidence with \eqref{eq:rag_case_score}--\eqref{eq:rag_retrieve_case}
            \STATE $A_{\mathrm{crit}}$ forms the reference levels using \eqref{eq:rag_case_weight}--\eqref{widehat phi}, and outputs $\left(a^{(n),r},\delta\boldsymbol d^{(n),r}\right)$ by \eqref{eq:critic_output_new}
            \IF{$a^{(n),r}=1$}
                \STATE Set $\boldsymbol c^{(\!n\!)} \!\leftarrow\! \boldsymbol c^{(\!n\!),r}$, send $\boldsymbol c^{(\!n\!)}$ to the lower level, \textbf{break}
            \ELSE
                \STATE $A_{\mathrm{plan}}$ revises the adjustment by \eqref{eq:planner_revise}
            \ENDIF
        \ENDFOR
        \IF{$\boldsymbol c^{(n)}$ is not assigned}
            \STATE Set $\boldsymbol c^{(\!n\!)} \leftarrow \boldsymbol c^{(\!n\!),R\!-\!1}$, and send $\boldsymbol c^{(\!n\!)}$ to the lower level
        \ENDIF
        \FOR{each slot $t\in\mathcal T^{(n)}$}
            \STATE With the estimated CSI $\widehat{\mathbf H}^{t}$, the lower level solves \eqref{eq:slot_energy_min_zf_wc} via CVX and obtains ${\bf W}^{t,*}\big(\boldsymbol c^{(n)}\big)$, and the slot-level feedback $\mathbf u^t$ is collected by the upper level
        \ENDFOR
        
        \STATE The upper level forms $\mathcal D^{(n)}$ and updates $\mathcal E^{(n+1)}$ with \eqref{eq:trace_def_multiagent} and \eqref{eq:experience_def_multiagent}, computes $\mathrm{KPI}^{(n)}$ and $G(\mathrm{KPI}^{(n)})$ with \eqref{eq:overall_bilevel_g} and \eqref{upper obj G}, and udpates the objective in \eqref{eq:overall_bilevel_a}
        \STATE Based on \eqref{eq:rag_case_memory}, the upper level computes $\Delta\boldsymbol c^{(n)}$, and the RAG module updates $\mathcal M_{\mathrm{case}}^{(n+1)}$
    \ENDFOR
    \STATE \textbf{Return:} $\{\!\boldsymbol c^{(\!n\!)},\mathrm{KPI}^{(\!n\!)}\!\}_{n=0}^{N\!-\!1}$, $\{\!{\bf W}^{t,*}\!\}_{t\!=\!1}^{T}$, $\sum_{n=0}^{N\!-\!1}\!\gamma^{(\!n\!)}\!G\!(\!\mathrm{KPI}^{(\!n\!)}\!)$
    \end{algorithmic}
\end{algorithm}

\subsection{Overall Algorithm Implementation}

\textbf{Algorithm~\ref{alg:agentic_ltpo}} summarizes the implementation of the proposed Agentic-LTPO framework. 
At each large timescale interval $n$, agent $A_{\mathrm{int}}$ first grounds the operator's policy input into the structured context and feasible decision domain, after which agent $A_{\mathrm{obs}}$ summarizes the current environment state. Conditioned on this environment state, the planner--critic coordination between agents $A_{\mathrm{plan}}$ and $A_{\mathrm{crit}}$ performs at most $R$ refinement rounds with RAG-assisted retrieval until a feasible configuration is accepted and passed to the lower level. The accepted configuration is fixed throughout interval $n$, while the fast solver computes the beamforming solutions based on the instantaneous CSI per slot. At the end of interval $n$, the realized lower-level beamforming solutions are aggregated to update the experience buffer, the designed KPIs, and the case memory for the next upper-level update. The algorithm can be initialized offline by a tuple that is consistent with the adopted configuration and memory settings.

\section{Experiments}


In this section, we evaluate the proposed Agentic-LTPO framework for a CF-MIMO system with $L=16$ distributed APs jointly serving $K = 8$ single-antenna users in a square service area of size $1~\mathrm{km}\times 1~\mathrm{km}$. Each AP is equipped with a uniform linear array (ULA) with $M=4$ transmit antennas and half-wavelength antenna spacing.
Unless otherwise specified, the default experiments use $L=16$ APs fixed at the grid points of a $4\times4$ layout spanning the $1~\mathrm{km}\times1~\mathrm{km}$ service area.
We set the pre-specified time $T=600$ time slots with $N=30$ large timescale intervals, each containing $T_s=20$ slots.
In each interval, the $K$ users are independently and uniformly dropped in the square service area. 
The user locations in each interval follow a homogeneous binomial point process in the area. The user locations are fixed within each large interval and regenerated independently across intervals.
To model the downlink channels, we adopt the standard large-scale/small-scale decomposition that is widely used in the CF-MIMO literature, e.g.,~\cite{ngo2017cellfree}. 
For each slot $t\in\mathcal{T}^{(n)}$, the channel from AP $\ell$ to user $k$ is generated as
\begin{align}
\mathbf{h}_{\ell k}^{t}
=
\sqrt{\beta_{\ell k}^{(n)}}\,\mathbf{g}_{\ell k}^{t},
\qquad
\mathbf{g}_{\ell k}^{t}\sim\mathcal{CN}\!\left(\mathbf{0},\mathbf{I}_M\right),
\end{align}
where 
$\mathbf{g}_{\ell k}^{t}$ denotes the small-scale Rayleigh fading vector, and $\beta_{\ell k}^{(n)}$ is the large-scale fading coefficient that remains unchanged within the $n$-th large timescale interval. The large-scale fading coefficient is modeled as
\begin{align}
\beta_{\ell k}^{(n)}
=
10^{\frac{\mathrm{PL}_{\ell k}^{(n)}+F_{\ell k}^{(n)}}{10}},
\end{align}
where $\mathrm{PL}_{\ell k}^{(n)}$ is the path loss (in dB) and $F_{\ell k}^{(n)}$ denotes the shadow fading (in dB). Considering an urban microcell setting, $\mathrm{PL}_{\ell k}^{(n)}[\mathrm{dB}] = -30.5-36.7\log_{10}\!({d_{\ell k}^{(n)}})$, where $d_{\ell k}^{(n)}$ is the distance between AP $\ell$ and user $k$ in interval $n$; the shadow fading yields $F_{\ell k}^{(n)}\sim\mathcal{N}(0,4^2)\ \mathrm{dB}$.
Unless otherwise specified, $\beta_{\ell k}^{(n)}$ is unchanged within any interval $n$ and updated independently across intervals, and $\mathbf{g}_{\ell k}^{t}$ is independently generated across slots.
The CSI uncertainty is set as $\delta_{\ell k}^{t} \!=\! \epsilon\|\hat{\mathbf{h}}_{\ell k}^{t}\|_2, \forall \ell, k, t$ with the default uncertainty level $\epsilon \!=\! 0.05$.

For the lower-level beamforming design, the noise variance is $(\sigma_k^t)^2=-167~\mathrm{dBm/Hz}$, $\forall k\in\mathcal{K}$. All APs are assigned the same nominal per-AP power budget $P_{\ell}^{\max}=-50~\mathrm{dBm/Hz}$, $\forall \ell\in\mathcal{L}$. The effective per-AP power limit in interval $n$ is adjusted by the upper-level coefficient $\phi_{\ell}^{(n)}$.

For the upper-level multi-agent setting, we manually generate $P^{(n)}$, which expresses only the operator's operational objectives and preferences in interval $n$ and does not directly specify the configuration values. The values of $R_{\mathrm{ref}}$, $\mathrm{EE}_{\mathrm{ref}}$, and $J_{\mathrm{ref}}$ are obtained offline with $6$ intervals, and the initial configuration $\boldsymbol{c}^{(-1)}$ in \textbf{Algorithm 1} is $\Gamma_k^{(-1)}=0~(\mathrm{dB}), \forall k \in\mathcal{K}$, and $\phi_\ell^{(-1)}=1, \forall \ell\in\mathcal{L}$. The empirically supported tolerance thresholds in \eqref{eq:critic_accept}, i.e., the admissible deviations of the candidate configuration components from their retrieved reference levels, are $\varepsilon_{\Gamma,k}^{(n),r} = 0.75+0.25r\ (\mathrm{dB})$ and $\varepsilon_{\phi,\ell}^{(n),r} = 0.05+0.03r$, respectively.
The other parameters associated with the system are listed in Table I.

\begin{table}[t]
\centering
\caption{simulation parameters not explicitly specified in the text.}
\label{tab:simulation_parameters}
\footnotesize
\setlength{\tabcolsep}{3pt}
\renewcommand{\arraystretch}{1.08}
\begin{tabular}{@{}p{0.46\columnwidth}p{0.48\columnwidth}@{}}
\toprule
\textbf{Parameter} & \textbf{Value} \\
\midrule



$R$,  & $3$ \\


$\lambda_R$, $\lambda_{\mathrm{EE}}$, $\lambda_J$ & $0.55$, $0.3$, $0.15$ \\

$R_{\mathrm{ref}}$, $\mathrm{EE}_{\mathrm{ref}}$, $J_{\mathrm{ref}}$
&
$1.25$, $7.0\times10^9$, $18.5$ \\

$|\mathcal{M}_{\mathrm{pol}}^{(0)}|$, $|\mathcal{M}_{\mathrm{case}}^{(0)}|$, $|\mathcal{E}^{(0)}|$
&
$3$, $6$, $6$ \\

$K_{\mathrm{pol}}$, $\tau_{\mathrm{pol}}$, $K_{\mathrm{case}}$, $\tau_{\mathrm{dir}}$
&
$3$, $0.85$, $3$, $0.5$ \\

$\lambda_{\mathrm{op}}$, $\lambda_{\mathrm{diag}}$, $\lambda_{\mathrm{dir}}$
&
$0.7$, $0.2$, $0.1$ \\

$\rho_{\Gamma}$, $\rho_{\phi}$ & $0.5$, $0.5$ \\

$(\alpha_{m,0},\alpha_{m,1},\alpha_{m,2})$, $\forall m$
&
$(1,2,0.01)$ \\

$(\tau_{\mathrm{in}},\tau_{\mathrm{out}})_{A_{\mathrm{int}}}$, $(\tau_{\mathrm{in}},\tau_{\mathrm{out}})_{A_{\mathrm{obs}}}$, 
& $(6,2)$, $(8,2)$ \\

$(\tau_{\mathrm{in}},\tau_{\mathrm{out}})_{A_{\mathrm{plan}}}$, $(\tau_{\mathrm{in}},\tau_{\mathrm{out}})_{A_{\mathrm{crit}}}$
& $(9,2)$, $(7,2)$ \\ 
\bottomrule
\end{tabular}
\end{table}
Since Agentic-LTPO jointly considers slow-timescale policy interpretation, cross-timescale configuration adaptation, and slot-level CF-MIMO beamforming, no existing methods are directly comparable. 
Particularly, classical beamforming methods, e.g.,~\cite{shi2011wmmse}, typically assume a fixed problem formulation, while learning-based controllers, e.g.,~\cite{cai2024deep,wang2024joint}, have been designed under pre-specified rewards or optimization templates. 
For the language-grounding study, we further consider an \emph{oracle structured-policy} setting, where the natural language policy is replaced by its manually specified structured counterpart. This setting is not a standalone benchmark; rather, it separates the effect of policy interpretation from that of the decision logic.
We consider the following benchmarks for Agentic-LTPO:
\begin{itemize}
    \item \textbf{Static Configuration}: A pre-tuned upper-level configuration $\boldsymbol{c}$ remains fixed throughout all periods.
    
    \item \textbf{Single-Agent Controller}: A single LLM-based controller directly maps the current policy input and environment summary to the next configuration to generate $\boldsymbol{c}^{(n)}$.
\end{itemize}
We also test the following ablated versions of Agentic-LTPO:
\begin{itemize}

    \item \textbf{Agentic-LTPO without RAG}: The proposed framework with the RAG module removed.
    
    \item \textbf{Agentic-LTPO without $A_{\mathrm{crit}}$}: The proposed framework with the performance critic and the corresponding refinement loop deactivated.
\end{itemize}

We consider two policy-generation settings in the experiments. In the first setting, the operator's policy is independently sampled for each large timescale interval, so that the policy tendency may change arbitrarily across adjacent intervals. This setting evaluates the adaptability of the upper-level controller under rapidly varying policy inputs. In the second setting, the policy sequence is generated in a piecewise-stationary manner, where every ten intervals share the same dominant policy tendency, e.g., throughput-oriented, QoS-oriented, or energy-saving operation. This setting evaluates whether the upper level can reliably track policy regimes and adjust the lower-level problem configuration accordingly.

\begin{figure}[!t]
	\centering
	\subfloat[{\centering Random policy setting.}]{
		\includegraphics[width=1.65in]{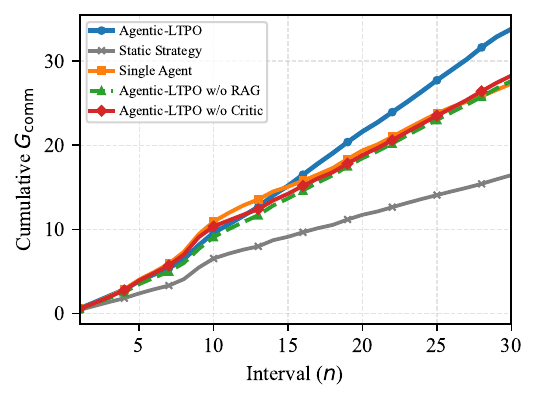}
		\label{fig:cum_utility_random}}
	\subfloat[{\centering Piecewise-stationary policy setting.}]{
		\includegraphics[width=1.65in]{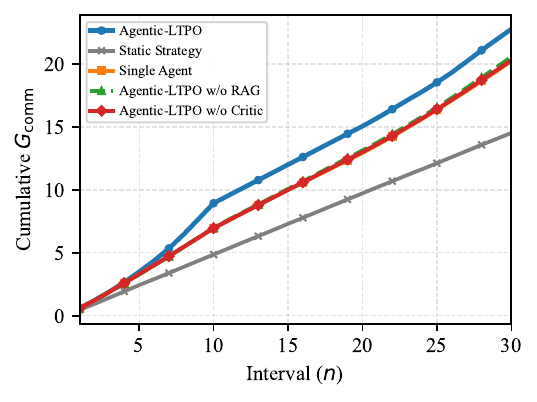}
		\label{fig:cum_utility_piecewise}}
	\caption{Cumulative communication utility of different upper-level controllers under two policy-generation settings.}
	\label{fig:cum_utility}
\end{figure}
We first evaluate the communication utility metric $G_{\mathrm{comm}}^{(n)} \triangleq
\frac{\lambda_R}{T_s}\sum_{t\in\mathcal{T}^{(n)}}\frac{R^t}{R_{\mathrm{ref}}} + \frac{\lambda_{\mathrm{EE}}}{T_s}\sum_{t\in\mathcal{T}^{(n)}}\frac{\mathrm{EE}^t}{\mathrm{EE}_{\mathrm{ref}}}$, which removes the upper-level inference energy from $G(\mathrm{KPI}^{(n)})$ in \eqref{upper obj G}. 
Fig.~\ref{fig:cum_utility} plots $\sum_{i=1}^{n} G_{\mathrm{comm}}^{(i)}$ of different benchmarks under both random and piecewise-stationary policy settings. Agentic-LTPO yields the largest cumulative communication-utility gain in both settings, showing that it can more effectively translate operator intents into suitable configurations across intervals. By contrast, the static configuration strategy remains fixed and therefore accumulates a significant mismatch as the policy evolves across intervals. 
Removing either the RAG module or the critic weakens the long-term gain, indicating that both retrieved evidence and planner--critic refinement contribute to sustained policy adaptation. The single-agent controller outperforms the static baseline, but is worse than the proposed design; simply introducing an LLM is insufficient for reliable long-term configuration.

\begin{figure}[!t]
	\centering
	\subfloat[{\centering Normalized sum-rate component.}]{
		\includegraphics[width=1.65in]{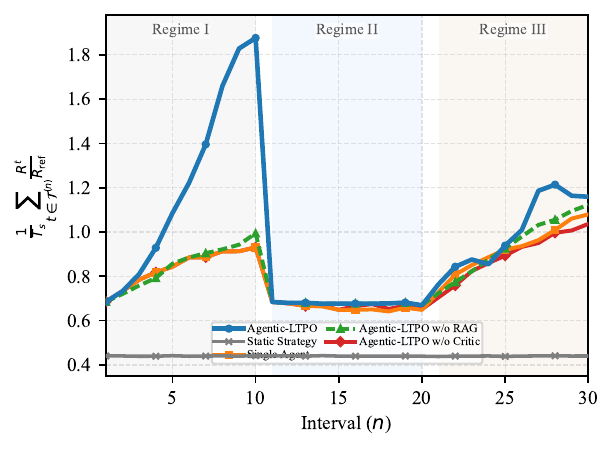}
		\label{fig:policy_following_a}}
	\subfloat[{\centering Normalized energy-efficiency component.}]{
		\includegraphics[width=1.65in]{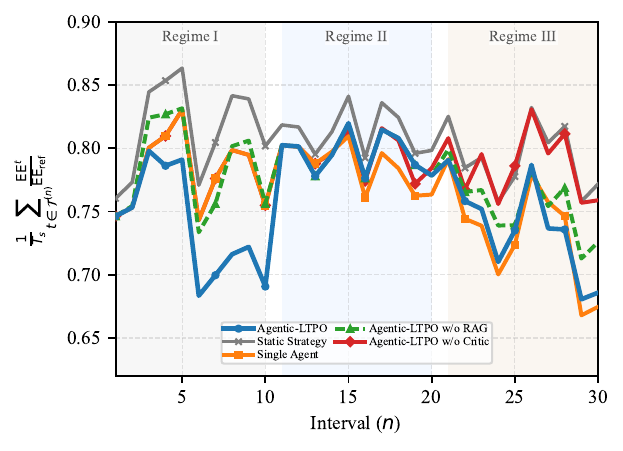}
		\label{fig:policy_following_b}}
	\caption{Policy-following KPI responses of different upper-level controllers under sustained operator policy regimes.}
	\label{fig:policy_following}
\end{figure}

Fig.~\ref{fig:policy_following} examines whether the upper-level controller follows the active operator policy at the level of individual KPI components. Unlike Fig.~\ref{fig:cum_utility}, which reports the cumulative communication utility, Fig.~\ref{fig:policy_following} plots the two normalized terms that constitute $G_{\mathrm{comm}}^{(n)}$, namely $\frac{1}{T_s}\sum_{t\in\mathcal{T}^{(n)}}\frac{R^t}{R_{\mathrm{ref}}}$ and $\frac{1}{T_s}\sum_{t\in\mathcal{T}^{(n)}}\frac{\mathrm{EE}^t}{\mathrm{EE}_{\mathrm{ref}}}$. The piecewise-stationary policy sequence is divided into three regimes, where Regimes I and III emphasize communication-oriented operation, while Regime II corresponds to an energy-saving operating tendency.

As shown in Fig.~\ref{fig:policy_following}(a), Agentic-LTPO increases the normalized sum-rate more effectively in the communication-oriented regimes. In Fig.~\ref{fig:policy_following}(b), the energy-efficiency is maintained at a high level during the energy-saving regime among the adaptive controllers. The static strategy also yields high energy efficiency because it is conservative, but it does not provide the corresponding sum-rate response in Regimes I and III. Hence, the result should be interpreted as a policy-following behavior rather than a single-metric dominance claim: Agentic-LTPO changes the dominant KPI response according to the active regime, whereas the single-agent controller and the ablated variants show weaker or less coordinated responses across the two KPI dimensions. 
Agentic-LTPO translates sustained policy regimes into both interpretable configuration changes and corresponding KPI-level responses.

\begin{figure}[!t]
	\centering
	\includegraphics[width=0.95\linewidth]{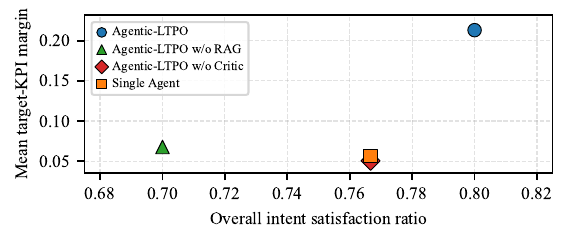}
	\caption{Policy-compliance operating points of different adaptive upper-level controllers.}
	\label{fig:intent_satisfaction}
\end{figure}

To quantify the above policy-complying behavior, Fig.~\ref{fig:intent_satisfaction} summarizes the intent satisfaction ratio and the corresponding target-KPI margin over the sustained policy sequence. For each large interval, the target KPI is selected according to the active policy regime: The normalized sum-rate metric is used in the throughput-oriented regimes, whereas the normalized energy-efficiency metric is used in the energy-saving regime. An interval is counted as intent-satisfied if the selected target KPI exceeds its regime-specific threshold. The target-KPI margin is the difference between the selected normalized KPI and the corresponding threshold, averaged over all intervals. The thresholds are used only to operationalize the natural-language policy into measurable compliance events, rather than to define a new optimization objective.

As shown in Fig.~\ref{fig:intent_satisfaction}, Agentic-LTPO lies in the upper-right region, indicating that it achieves both a higher overall intent satisfaction ratio and a larger positive target-KPI margin. The single-agent controller and the ablated variants can satisfy part of the policy sequence, but their margins are smaller, which implies weaker headroom relative to the active target KPI. Complementing the interval-wise KPI response in Fig.~\ref{fig:policy_following}, Fig.~\ref{fig:intent_satisfaction} provides a compact compliance-level summary of whether the response is aligned with the active operator intent.

\begin{figure}[!t]
	\centering
	\subfloat[{\centering Final utility--overhead tradeoff.}]{
		\includegraphics[width=1.65in]{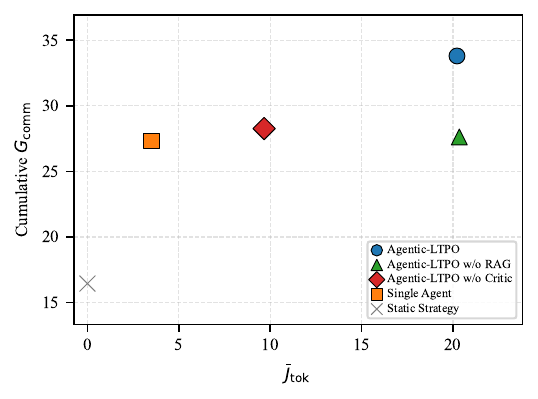}
		\label{fig:token_overhead_a}}
	\subfloat[{\centering Cumulative utility--overhead trajectory.}]{
		\includegraphics[width=1.65in]{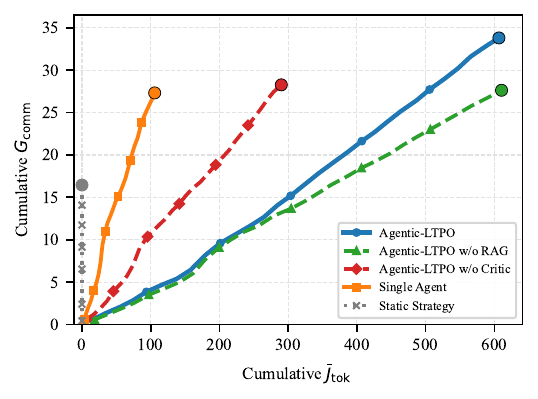}
		\label{fig:token_overhead_b}}
	\caption{Token-based upper-level reasoning overhead under the random policy setting.}
	\label{fig:token_overhead}
\end{figure}

Fig.~\ref{fig:token_overhead} reports the token-side reasoning overhead associated with the random-policy experiment in Fig.~\ref{fig:cum_utility}(a). The $x$-axis uses the token-based proxy $\bar{J}_{\mathrm{tok}}$, computed from the recorded input and output token counts of the upper-level controller; the $y$-axis keeps the communication-side metric $G_{\mathrm{comm}}^{(n)}$ unchanged. As shown in Fig.~\ref{fig:token_overhead}(a), Agentic-LTPO achieves the largest cumulative $G_{\mathrm{comm}}$ but also incurs a larger token cost than the single-agent controller and the static strategy. This is expected though, since the proposed architecture invokes policy interpretation, retrieval-assisted planning, and critic-based verification. Agentic-LTPO outperforms Agentic-LTPO without RAG even though their token-cost proxies are comparable, indicating that the communication-utility gain is not merely caused by more tokens. Fig.~\ref{fig:token_overhead}(b) shows the same tradeoff in the entire simulation horizon: The proposed method accumulates higher reasoning overhead as the intervals proceed; the additional cost contributes to a higher final communication utility. Agentic-LTPO exhibits a utility--overhead tradeoff: The multi-agent reasoning process is more expensive than a monolithic or static controller, but yields more effective long-term configuration under varying operator policies.

\begin{figure}[!t]
	\centering
	\includegraphics[width=0.95\linewidth]{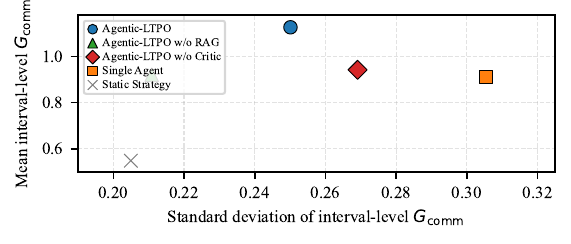}
	\caption{Mean--variability tradeoff of interval-level communication utility under the random policy setting.}
	\label{fig:mean_variability}
\end{figure}

To further examine whether the long-term gain in Fig.~\ref{fig:cum_utility}(a) is obtained through unstable interval-level behavior, Fig.~\ref{fig:mean_variability} plots the mean of $G_{\mathrm{comm}}^{(n)}$ against its standard deviation across intervals under the random policy setting. Agentic-LTPO attains the largest mean interval-level communication utility, while its variability remains lower than that of the single-agent controller and Agentic-LTPO without the critic. Agentic-LTPO without RAG has a slightly smaller standard deviation and produces a much lower mean utility, suggesting less responsive update behavior rather than a better operating point. The proposed design improves the mean--variability tradeoff: Retrieval-assisted grounding and critic-guided verification allow the upper level to make more effective adaptations, while avoiding larger fluctuations observed in the monolithic single-agent controller and the no-critic variant.

\begin{figure}[!t]
	\centering
	\subfloat[{\centering Average target robust QoS level.}]{
		\includegraphics[width=1.65in]{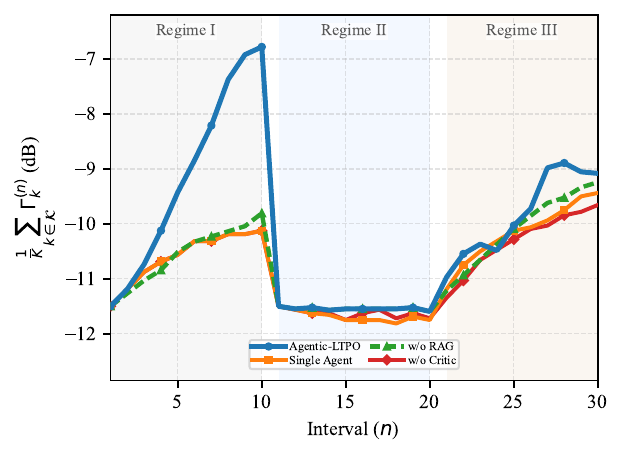}
		\label{fig:config_traj_a}}
	\subfloat[{\centering Average AP power-budget exposure.}]{
		\includegraphics[width=1.65in]{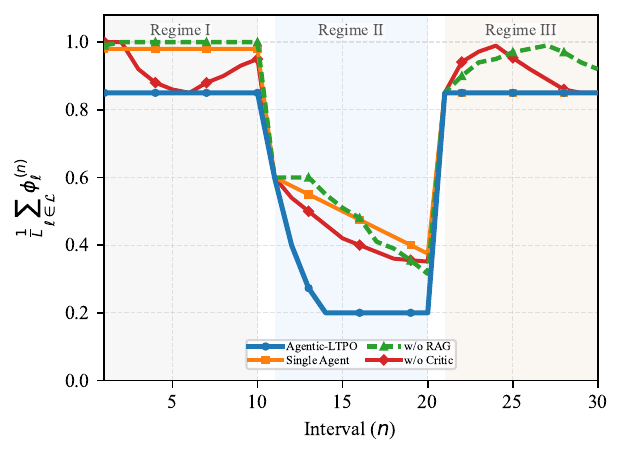}
		\label{fig:config_traj_b}}
	\caption{Configuration trajectories generated by different upper-level controllers under sustained operator policy regimes.}
	\label{fig:config_traj}
\end{figure}

To understand the cause of the above-mentioned cumulative communication-utility gain, Fig.~\ref{fig:config_traj} plots the interval-level configuration variables generated by the upper-level controllers, including the average target robust QoS level $\frac{1}{K}\sum_{k\in\mathcal{K}}\Gamma_k^{(n)}$ and the average AP power-budget exposure factor $\frac{1}{L}\sum_{\ell\in\mathcal{L}}\phi_\ell^{(n)}$. These curves provide a direct view of how policy semantics are translated into lower-level optimization parameters. In Regime I, Agentic-LTPO increases the average target robust QoS more aggressively, which is consistent with the throughput-oriented policy intent. In Regime II, it sharply relaxes the AP power budget exposure factor, thereby enforcing a more conservative power-allocation regime under the energy-saving intent. When the throughput-oriented policy returns in Regime III, both configuration variables are restored toward a more communication-favorable operating point. The single-agent controller and the ablated variants also exhibit partial adaptation, but their configuration changes are less responsive and less coordinated across the QoS and AP-exposure dimensions. This observation supports the role of retrieval-assisted grounding and critic-guided refinement in producing policy-consistent upper-level configurations.

\begin{figure}[!t]
	\centering
	\subfloat[{\centering Cumulative communication utility.}]{
		\includegraphics[width=1.65in]{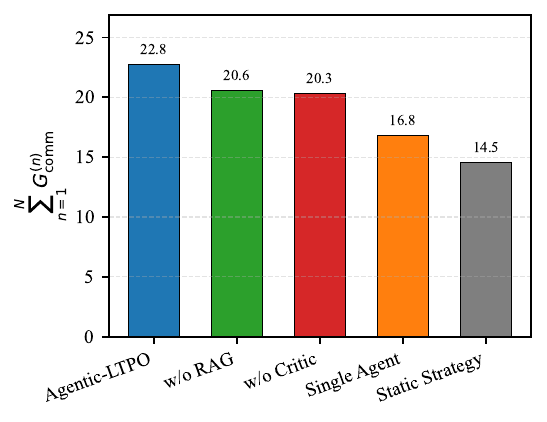}
		\label{fig:ablation_a}}
	\subfloat[{\centering Regime-wise average communication utility.}]{
		\includegraphics[width=1.65in]{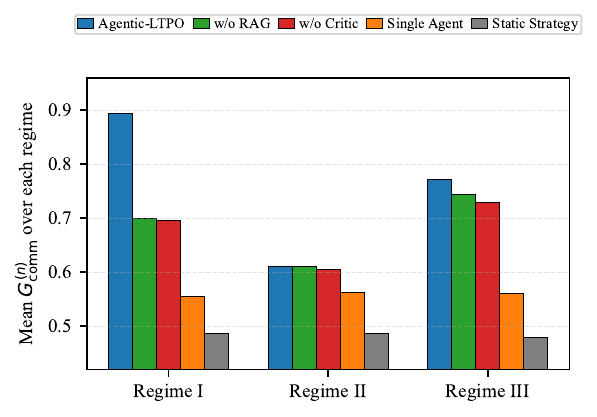}
		\label{fig:ablation_b}}
	\caption{Ablation study on the contributions of retrieval-assisted grounding and critic-guided refinement.}
	\label{fig:ablation}
\end{figure}

Fig.~\ref{fig:ablation} evaluates the contribution of the main upper-level components. Fig.~\ref{fig:ablation}(a) reports the cumulative communication utility $\sum_{n=1}^{N}G_{\mathrm{comm}}^{(n)}$; Fig.~\ref{fig:ablation}(b) reports the average $G_{\mathrm{comm}}^{(n)}$ within each sustained policy regime. Compared with the single-agent controller, Agentic-LTPO obtains a larger cumulative gain, showing that simply mapping the policy and environment summary to a configuration is insufficient for effective long-term adaptation. Removing the RAG module or the critic also degrades the cumulative utility, indicating that both historical evidence reuse and planner--critic verification contribute to the communication utility. The regime-wise comparison further shows that the advantage is the most visible in Regimes I and III, where the policy emphasizes communication-oriented operation and the upper level must actively restore a more aggressive configuration. In Regime II, the utility gap is smaller because all adaptive controllers move toward a more conservative configuration under the energy-saving intent. These results are consistent with the configuration trajectories in Fig.~\ref{fig:config_traj} and support the role of the proposed multi-agent structure in coordinating policy interpretation, retrieval grounding, and critic-guided refinement.

\begin{figure}[!t]
	\centering
	\subfloat[{\centering Cumulative communication utility.}]{
		\includegraphics[width=1.65in]{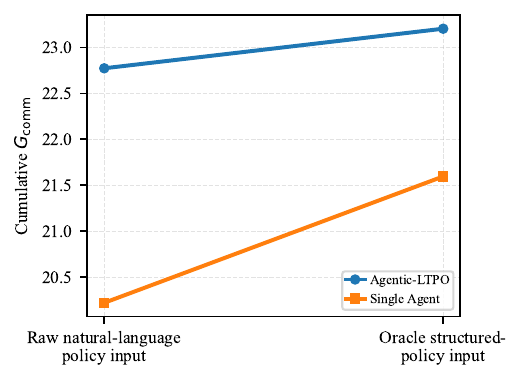}
		\label{fig:raw_vs_oracle_a}}
	\subfloat[{\centering Mean target-KPI margin.}]{
		\includegraphics[width=1.65in]{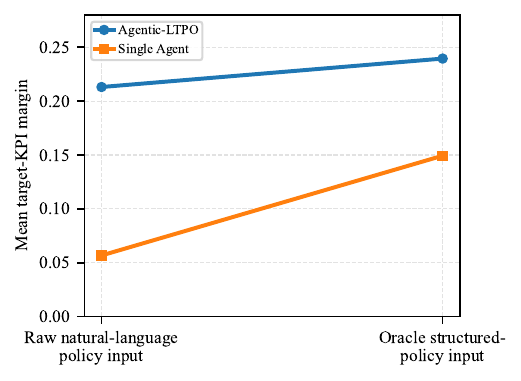}
		\label{fig:raw_vs_oracle_b}}
	\caption{Impact of raw natural-language and oracle structured-policy inputs on upper-level policy adaptation.}
	\label{fig:raw_vs_oracle}
\end{figure}
Fig.~\ref{fig:raw_vs_oracle} compares the cumulative communication utility and target-KPI margin under raw natural-language policy input and oracle structured-policy input. Since the oracle structured-policy input removes the ambiguity of natural language while keeping the downstream controller unchanged, the gap between the two input formats reflects the sensitivity of the controller to language grounding. As shown in Fig.~\ref{fig:raw_vs_oracle}(a), Agentic-LTPO maintains a higher cumulative communication utility than the single-agent controller under both input formats. 
Replacing the raw input with the oracle structured-policy input slightly increases the cumulative utility of Agentic-LTPO, with only a $1.9\%$ gain over the raw-input case, while the corresponding gain for the single-agent controller is $6.8\%$.
This indicates that the monolithic controller is more sensitive to the input representation. Fig.~\ref{fig:raw_vs_oracle}(b) shows the same trend from the policy-compliance perspective: The oracle structured-policy input increases the mean target-KPI margin for both controllers, while Agentic-LTPO retains a larger positive margin. In this sense, the gain of Agentic-LTPO is not solely due to stronger text processing; it also comes from the structured grounding and multi-agent decision flow.

\begin{figure}[htbp]
	\centering
	\subfloat[{\centering Cumulative communication utility.}]{
		\includegraphics[width=1.65in]{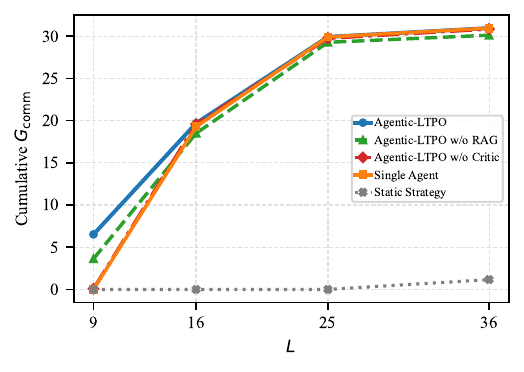}
		\label{fig:service_load_l_sweep_a}}
	\subfloat[{\centering QoS violation ratio.}]{
		\includegraphics[width=1.65in]{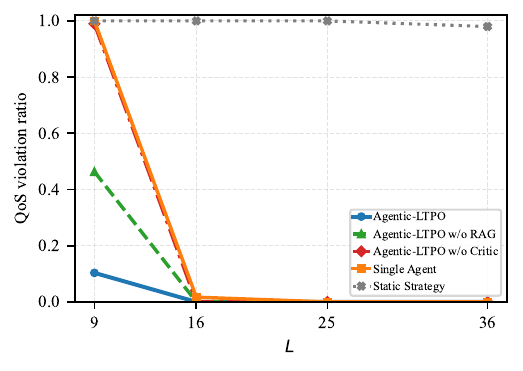}
		\label{fig:service_load_l_sweep_b}}
	\caption{Service-load robustness under different AP densities.}
	\label{fig:service_load_l_sweep}
\end{figure}

Fig.~\ref{fig:service_load_l_sweep} evaluates the behavior of different upper-level controllers under a service-load stress test by varying the number of APs, $L$, while keeping the user load fixed. Fig.~\ref{fig:service_load_l_sweep}(a) reports the accumulated communication utility $\sum_{n=1}^{N}G_{\mathrm{comm}}^{(n)}$; Fig.~\ref{fig:service_load_l_sweep}(b) reports the QoS violation ratio, defined as the proportion of user-slot instances in which the achieved robust QoS lower bound is below the configured target. When $L$ is small, the lower-level feasible region is more restrictive and the quality of the upper-level configuration is more important. In this regime, Agentic-LTPO achieves a larger cumulative $G_{\mathrm{comm}}$ and a lower QoS violation ratio than the ablated and single-agent controllers, indicating that the proposed retrieval-assisted and critic-guided decision flow produces more reliable lower-level configurations under sparser deployments of fewer APs.
As $L$ increases, the additional AP resources enlarge the feasible solution region and it becomes easier for all adaptive controllers to satisfy the QoS targets, hence reducing the violation ratio and narrowing the performance gap. The static strategy remains weak since it cannot adjust the lower-level configuration to the stressed operating condition.

\section{Conclusion}
In this paper, we presented a new agentic AI framework, named Agentic-LTPO, to tackle long-term performance optimization for wireless physical layer tasks. 
By decoupling agentic AI decisions from instantaneous optimization, we developed a nested bilevel framework to handle dynamic operator policies, together with stringent real-time constraints. 
To illustrate the advantages of Agentic-LTPO, we considered a CF-MIMO beamforming scenario, where an upper-level multi-agent system parameterizing the operator's changing policies and KPIs was designed to adapt the lower-level beamforming problem configuration based on policy inputs, environment summaries, and historical experience. 
Numerical results demonstrated that Agentic-LTPO improves cumulative communication utility by $57.2\%$ over the static baseline, while limiting the cumulative-utility gap between raw natural-language and oracle structured-policy inputs to $1.9\%$.

\bibliographystyle{IEEEtran}
\bibliography{references}

\end{document}